%% file: bare_jrnl_new_sample4.tex
\documentclass[lettersize,journal]{IEEEtran}
\usepackage{amsmath,amsfonts}
\usepackage{algorithm}
\usepackage{array}
\usepackage[caption=false,font=normalsize,labelfont=sf,textfont=sf]{subfig}
\usepackage{textcomp}
\usepackage{stfloats}
\usepackage{url}
\usepackage{verbatim}
\usepackage{graphicx}
\usepackage{cite}
\usepackage{threeparttable}
\usepackage{graphicx}
\usepackage{booktabs}
\usepackage{algpseudocode}
\usepackage{xcolor}
\usepackage{multirow}

\usepackage[most]{tcolorbox}
\usepackage{xcolor}
\usepackage{makecell}
\definecolor{mainbg}{RGB}{238,244,249}
\definecolor{mainframe}{RGB}{92,120,145}

\definecolor{databg}{RGB}{239,246,241}
\definecolor{dataframe}{RGB}{91,127,103}

\definecolor{visionbg}{RGB}{249,244,236}
\definecolor{visionframe}{RGB}{151,119,82}

\definecolor{criticbg}{RGB}{244,241,248}
\definecolor{criticframe}{RGB}{119,101,139}
\begin{document}

\title{RingMoClaw: An Experience-Inspired Multi-Agent Framework for Self-Evolving Research in Remote Sensing}

\author{
Kaiyue~Kang, Qixuan~He, Peijin~Wang, Yingchao~Feng, Chao~Ren, Kangxin~Wang, Wenhui~Diao, Yixiao~Wang, Liangjin~Zhao, Kaiwen~Wei, Nayu~Liu, Xian~Sun
\thanks{ K. Kang, Q. He and P. Wang contribute equally to this work. This work was supported by the National Natural Science Foundation of China under Grant 62425115, and partially supported by China Science and Technology Cloud (CSTCloud). (Corresponding author: Xian~Sun)}
\thanks{ K. Kang, Q. He, P. Wang, K. Wang, W. Diao and X. Sun are with the Aerospace Information Research Institute, Chinese Academy of Sciences, Beijing 100190, China, also with the School of Electronic, Electrical and Communication Engineering, University of Chinese Academy of Sciences, Beijing 100190, China, also with the University of Chinese Academy of Sciences, Beijing 100190, China, and also with the National Key Laboratory of Target Cognition and Application Technology (TCAT), Aerospace Information Research Institute, Chinese Academy of Sciences, Beijing 100190, China.}
\thanks{Y. Feng, C. Ren, Y. Wang, L. Zhao, K. Wei and N. Liu are with the Aerospace Information Research Institute, Chinese Academy of Sciences, Beijing 100190, China, and also with the National Key Laboratory of Target Cognition and Application Technology (TCAT), Aerospace Information Research Institute, Chinese Academy of Sciences, Beijing 100190, China.}
}


\maketitle

\begin{abstract}

\textcolor{black}{Remote sensing visual models have continuously advanced various interpretation tasks. However, the research process behind model improvement still heavily relies on manual expertise, requiring extensive trial-and-error iterations in model design, data processing, and performance diagnosis. Existing agent-based approaches mainly focus on task execution and workflow orchestration, while lacking the capability of autonomous research iteration for continuous performance optimization. To address this issue, we propose \textbf{RingMoClaw}, an experience-inspired self-evolving multi-agent framework for remote sensing visual interpretation. RingMoClaw integrates a research branch, a quality-control branch, and a dual-stream dynamic experience bus to establish a closed-loop optimization process covering strategy generation, experiment execution, independent review, and experience accumulation. The heterogeneous Critic mechanism provides stage-wise diagnosis and feedback, while the dual-stream experience bus incorporates external knowledge and internal experimental experience to guide strategy evolution and eliminate ineffective searches. Extensive experiments on four remote sensing downstream tasks, including object detection, scene classification, semantic segmentation, and change detection, demonstrate the effectiveness and generalization of RingMoClaw. Compared with the corresponding baseline models, RingMoClaw
improves performance by 1.84\% mAP$_{50}$ on object detection and achieves
consistent gains across the other three tasks, while reducing the required
evolution steps by over 40\% compared with existing research automation
frameworks. These results suggest that RingMoClaw offers a feasible route from
task execution toward continuous research driven model evolution in remote
sensing.}

\end{abstract}



\section{Introduction}
\label{sec:introduction}

Remote sensing earth observation is an essential means of acquiring dynamic information about the land surface, and provides critical geospatial support for environmental monitoring, disaster response, urban planning, and agricultural management~\cite{ref1,ref2}. With the continuous improvement of remote sensing observation capability, the scale of remote sensing data keeps growing and the information they contain becomes increasingly rich. This places higher demands on the efficiency and accuracy of remote sensing interpretation. In recent years, remote sensing visual interpretation methods have made steady progress in tasks such as object detection, semantic segmentation,  scene classification and land-cover mapping~\cite{cheng2017remote, li2020object,sun2022ringmo,ma2021factseg,zheng2023farseg++, mei2024comprehensive,yu2025glc}. 
\textcolor{black}{Beyond network architecture design, recent studies pay more attention to learning strategies that cope with limited annotation and distribution shift, including domain adaptation, few shot learning, and semi
supervised learning for scene classification~\cite{jose2026unsupervised,QIU2024368,FENG2025104335}, as well as few shot learning, active label selection, and cross domain consistency for object detection~\cite{yang2024few,pan2025locate,ALS,Dualteacher}. These results indicate that accuracy gains depend on coordinated choices across model architecture, data processing, and training strategy.}
However, the underlying research process still relies heavily on manual expertise. For a given task, researchers often need to repeatedly adjust model design, training strategy, data processing, and result analysis through trial and error. This expert-driven paradigm is costly and time-consuming, and also limits the reproducibility, scalability, and long-term optimization efficiency of remote sensing algorithm development.
\begin{figure*}[!t]
    \centering
    \includegraphics[width=0.9\textwidth]{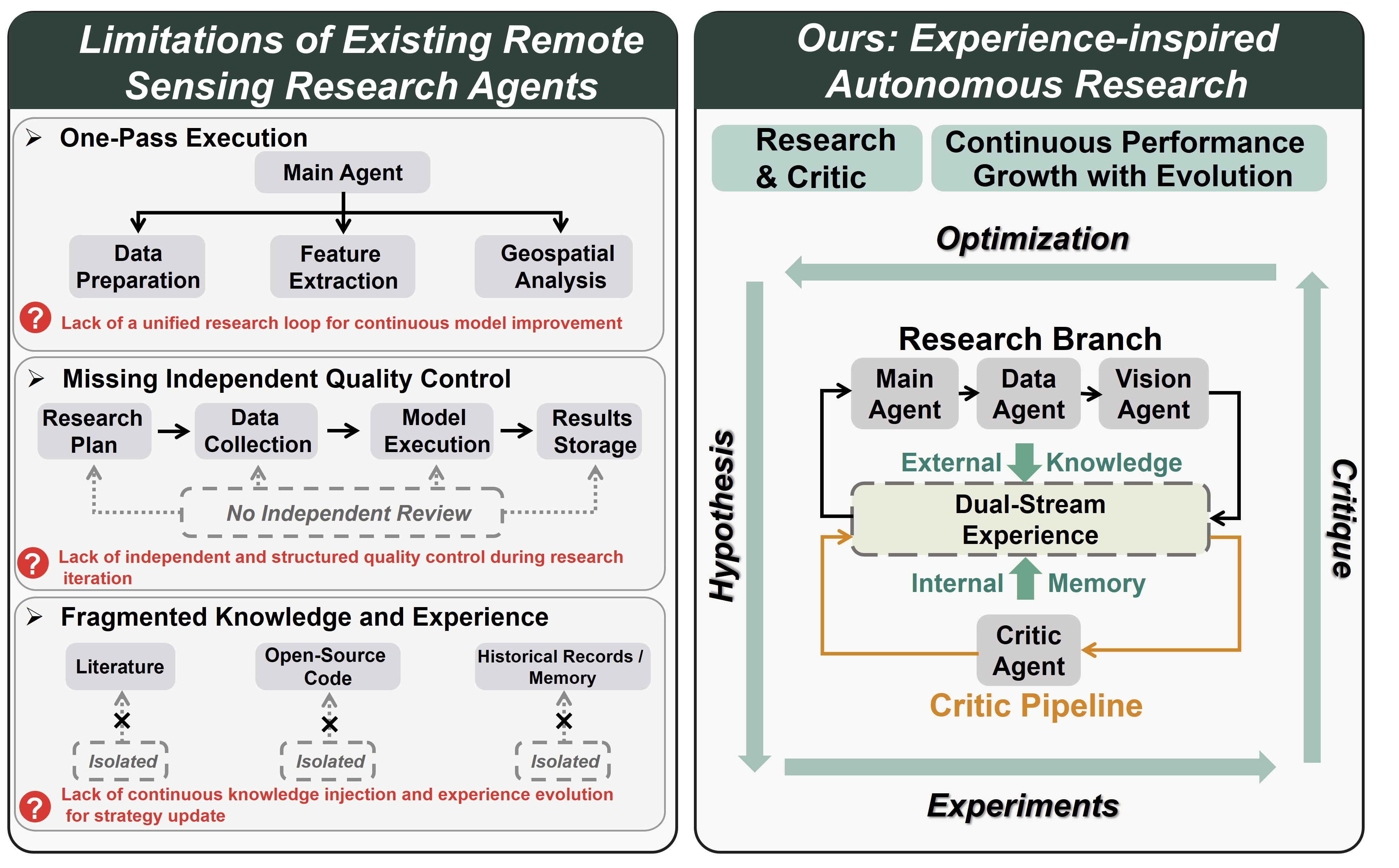}
    \caption{Comparison of existing remote sensing research agents and RingMoClaw. RingMoClaw introduces a unified research loop with independent critique and dual-stream experience evolution for continuous model improvement.}
    \label{fig:paradigm}
\end{figure*}

The capabilities of large language models (LLMs) in complex reasoning, task planning, and code generation~\cite{ref8,ref9} have accelerated the development of agent technology in geospatial and remote sensing domains. Existing studies can be broadly grouped into two categories. The first category focuses on tool calling and workflow orchestration. Representative systems such as GeoGPT~\cite{ref10}, ChangeGPT~\cite{ref11}, GeoLLM~\cite{manvi2024geollm}, and GeoColab~\cite{ref13} extend the scope of agents in geospatial task understanding, remote sensing change analysis, multi agent workflow collaboration, and geospatial code generation, and improve the automated execution of complex geospatial tasks through natural language understanding, task decomposition, and module coordination. The second category focuses on unified agent organization and open-ended capability extension for complex remote sensing scenarios. Earth-Agent~\cite{ref14} and CangLing-KnowFlow~\cite{chen2025cangling} contribute to cross modal earth observation reasoning, procedural knowledge organization, evolving memory, and multi step workflow management, and strengthen the reasoning and scheduling capability of agents on complex remote sensing tasks. Built upon the OpenClaw research agent paradigm~\cite{OpenClawRL2026}, OpenEarth-Agent~\cite{ref16} extends this line from tool calling to tool creation and enables agents to dynamically generate dedicated tools for unseen data and tasks, which improves the autonomous execution of earth observation tasks in open environments. Overall, these studies have substantially advanced the automation of task execution in remote sensing and geospatial analysis, and have pushed agents from calling existing capabilities toward organizing and extending the task solving workflow.

Although the above studies have substantially advanced the automation of task execution in remote sensing and geospatial analysis, the main focus remains on task solving, workflow orchestration, and capability extension. Support for research iteration aimed at continuous performance improvement is still limited. This limitation is more evident in remote sensing vision tasks, which often involve long-tailed category distributions, missed detection of small objects, and domain-specific prior constraints. As shown in Fig.~\ref{fig:paradigm}, existing remote sensing agents still have the following three limitations:

\begin{itemize}
    \item \textbf{Lack of a unified research loop for continuous model improvement.} Existing systems can decompose tasks, call tools, and organize workflows for a given objective~\cite{ref10,ref11,manvi2024geollm,ref13}. However, most systems are designed for one-pass task completion. A closed loop that connects hypothesis generation, experiment execution, performance analysis, and strategy update is still missing. As a result, sustained model optimization remains difficult.
    
    \item \textbf{Lack of independent and structured quality control during research iteration.} Some recent frameworks improve reasoning and workflow management for complex tasks~\cite{ref14,ref15}. However, explicit mechanisms for independent review are still limited. Structured diagnosis of plan design, data processing, and experimental results is usually absent. Quantitative feedback for key bottlenecks is also insufficient. This makes targeted correction during iteration difficult.
    
    \item \textbf{Lack of continuous knowledge injection and experience evolution for strategy update.} Existing systems have explored retrieval, memory, and even tool creation in open environments~\cite{ref14,ref15,ref16}. However, recent literature, open-source implementations, and historical experimental records are not jointly integrated into a unified optimization process. External knowledge is difficult to transform into actionable strategies. Internal experimental experience is also difficult to accumulate as reusable guidance. Consequently, dynamic expansion of the strategy space and pruning of inefficient paths remain limited.
\end{itemize}


To address the above limitations, this paper proposes \textbf{RingMoClaw}, an experience-inspired self-evolving multi-agent framework built upon RingMo~\cite{sun2022ringmo} for remote sensing visual tasks. RingMoClaw is designed to support not only task execution, but also iterative model optimization for continuous performance improvement. The framework consists of a research branch for plan generation, data processing, and experiment execution, a quality control branch for independent diagnosis and feedback, and a dual-stream dynamic experience bus for external knowledge injection and internal experience accumulation. Together, these components form a unified optimization loop for sustained research iteration in remote sensing visual interpretation. The main contributions of this paper are summarized as follows:

\begin{enumerate}

\item \textcolor{black}{We propose \textbf{RingMoClaw}, an experience-inspired self-evolving multi-agent framework for remote sensing visual tasks. Beyond conventional agent systems for task execution, RingMoClaw automates the research iteration behind model improvement, where the optimization trajectory is dynamically refined through experimental feedback and accumulated experience.}

\item \textcolor{black}{We design a heterogeneous \textbf{Critic mechanism} for structured quality control. By conducting staged review across the research pipeline, the Critic quantitatively evaluates task completion, bottleneck impact, and historical experience relevance. The resulting score guides experience selection and triggers pruning of unpromising paths.}

\item \textcolor{black}{We propose a \textbf{Dual-stream Dynamic Experience Bus} that integrates external knowledge flow and internal experimental flow into the research loop. Retrieved knowledge and experimental trajectories are transformed into reusable experience and mapped back to the structured search space, enabling dynamic expansion and refinement of optimization strategies during evolution.}

\item \textcolor{black}{We evaluate RingMoClaw on four representative remote sensing tasks, including object detection, scene classification, semantic segmentation, and change detection, to assess its capability in autonomous research iteration. The framework improves the corresponding baselines by 1.84\% mAP$_{50}$ on object detection and by up to 2.87\% on the remaining three tasks, and reaches convergence with at least 40\% fewer evolution rounds than existing research automation frameworks.}
    
\end{enumerate}

\section{Related Work}
\input{sections/related_work}

\section{Methodology}

\subsection{Overall Framework}
\label{sec:overall}


To move beyond the task-specific tool-calling paradigm toward genuine intelligent synthesis, we propose RingMoClaw, a multi-agent architecture for experience-inspired autonomous research evolution. As illustrated in Fig.~\ref{fig:overall_framework}, the architecture is composed of a Self-Evolving Research Pipeline, a Critic mechanism, and a Dual-stream Dynamic Experience Bus.

The Self-Evolving Research Pipeline serves as the operational core, where the Main Agent, Data Agent, and Vision Agent collaborate to decompose complex research objectives (e.g., fine-grained detection) into a staged evolution trajectory. Unlike conventional frameworks, this pipeline does not merely execute commands; it instantiates potential solutions from an evolving knowledge base.

The Critic mechanism provides the necessary independent external review. By conducting a three-phase diagnostic (Plan, Data, and Result), it identifies experimental anomalies and accuracy bottlenecks that often escape standard automated systems. These insights are quantified into a Multi-Dimensional Score $t$, ensuring that every failure or success is distilled into reusable intelligence.

The Dual-stream Dynamic Experience Bus acts as the system's ``cognitive center.'' It hosts a Dual-Stream Experience Engine that aggregates external RAG-based knowledge with internal experimental feedback. By mapping these streams onto a Structured Search Space $\mathcal{S}$, the bus continuously prunes unpromising paths and refines optimization directions. Through this mechanism, every iteration of the research branch is no longer a blind trial-and-error but a precise step informed by the quality control branch, driving the system to converge rapidly toward a task-specific improved solution.

\textcolor{black}{
For reproducibility, all agents are driven by fixed and version-controlled system prompts. 
Each prompt specifies the role-specific behavior, declared inputs, and a structured JSON output contract. 
At each evolution round, the agent receives a serialized state containing the current task, identified bottleneck, candidate under review, historical summary, and retrieved experience. 
All outputs are schema-validated before entering the subsequent stage. 
The complete prompts are provided in Appendix~\ref{app:agent_prompts}.
}

\begin{figure}[!t]
    \centering
    \includegraphics[width=0.5\textwidth]{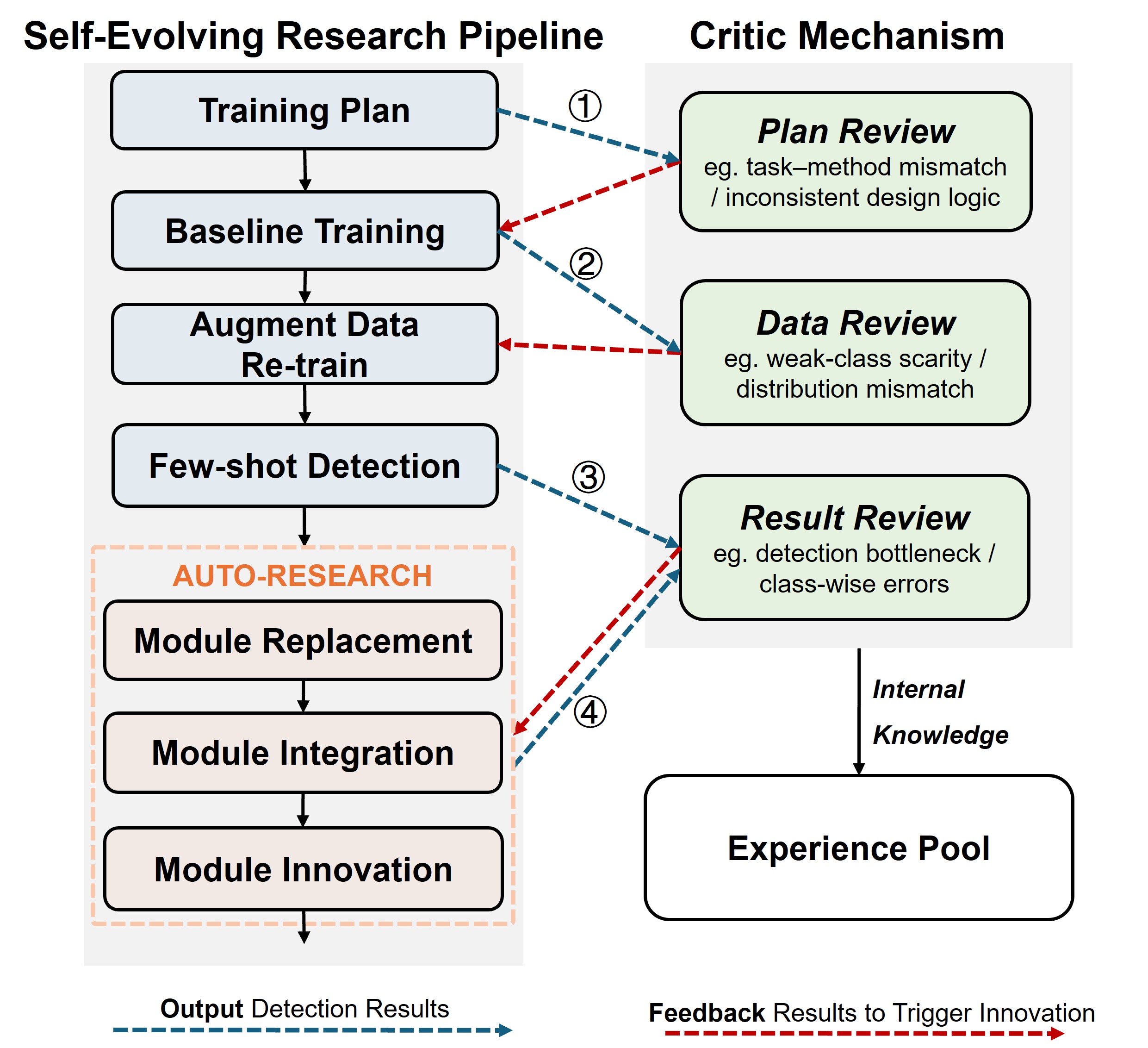}
\caption{Object-detection workflow of the proposed framework. The research pipeline performs staged model optimization, while the Critic mechanism reviews intermediate outputs and returns feedback for strategy revision and experience accumulation.}
    \label{fig:detection_pipeline}
\end{figure}

To illustrate this process, we take object detection as a representative example. As shown in Fig.~\ref{fig:detection_pipeline}, the self-evolving pipeline and the critic mechanism interact throughout the optimization loop. Algorithm~\ref{alg:detection_flow} records one representative trajectory that the system autonomously explored on FAIR1M. The specific techniques in it, such as ClassAwareSampler, FocalLoss~\cite{lin2017focal}, and BFPA~\cite{lin2025attention}, are outcomes of the search over $\mathcal{S}$ rather than a predefined script. Along this trajectory, the self-evolving pipeline performs staged optimizations spanning baseline training, data augmentation, few-shot detection, and automated module replacement, integration, and innovation. In parallel, the critic mechanism conducts multi-level reviews on plans, data, and results to guide iterative improvements. This collaborative mechanism ties model optimization closely to structured experience accumulation, converting experimental successes into permanent system knowledge.




\begin{algorithm}[t]
\caption{An Instantiated Self-Evolving Trajectory for Object Detection on FAIR1M}
\label{alg:detection_flow}
\begin{algorithmic}[1]
\Require Task $T$, initial model $M$, experience pool $\mathcal{E}$
\Ensure Optimized detector $M^*$
\Statex \textbf{Stage 1: Initialization and Diagnostic Planning}
\State Main Agent analyzes task $T$ and generates an initial optimization roadmap $\mathcal{R}$;
\State Vision Agent trains the baseline; Critic mechanism identifies weak-class bottlenecks;
\Statex \textbf{Stage 2: Iterative Policy Optimization}
\State Main Agent selects a data-level direction from $\mathcal{S}_{\text{data}}$ (here instantiated as augmentation: rotation, flip, mixup); retained if recall improves;
\State Main Agent selects a few-shot direction for weak classes from the current search space;
\If{overall accuracy decreases}
    \State Critic mechanism rejects the direction and records the failure in $\mathcal{E}$ to avoid repetition;
\EndIf
\Statex \textbf{Stage 3: Automated Module Evolution (Auto-Research)}
\State Main Agent selects a resampling strategy from $\mathcal{S}_{\text{data}}$ (here instantiated as ClassAwareSampler);
\State Main Agent integrates a training-strategy module and an externally retrieved neck module from $\mathcal{S}_{\mathrm{train}}$ and $\mathcal{S}_{\mathrm{arch}}$ (here instantiated as FocalLoss~\cite{lin2017focal} and the reused BFPA module~\cite{lin2025attention}), each promoted only after passing agile verification;
\State Main Agent proposes two newly designed modules beyond the existing search space (the Multi-Scale Feature Refinement and Contrastive Prototype Refinement modules), and the Critic mechanism validates their effect;
\State Validated knowledge is accumulated into $\mathcal{E}$ for cross-task evolution;
\State \Return $M^*$
\end{algorithmic}
\end{algorithm}

\subsection{Critic Mechanism: Heterogeneous Review and Targeted Feedback}
\label{sec:critic}

Within the self-evolving framework, the Critic mechanism is responsible for independent diagnosis, quantitative evaluation, and targeted feedback on the outputs of different stages in the execution chain. It serves as a key mechanism for sustaining model optimization during research iteration. Its design follows two principles: heterogeneity and functional separation.

At the model level, the execution chain is built on GLM-5 Turbo, while the Critic mechanism is powered by Minimax 2.7. The two models differ in pretraining data, reasoning style, and knowledge preference. This heterogeneous configuration allows the Critic mechanism to examine the outputs of the execution chain from a distinct perspective and reduces the homogeneous bias that may arise from self-evaluation by the same model. At the functional level, the Critic mechanism participates in neither plan generation nor experiment execution, and is only responsible for review and diagnosis. The execution chain is oriented toward task completion, whereas the Critic mechanism is oriented toward problem identification and quality assessment. This functional separation provides an external safeguard for the research loop and helps the system avoid ineffective optimization paths.

\subsubsection{Staged Review across the Full Chain}
\label{sec:critic_review}

\begin{figure}[!t]
    \centering
    \includegraphics[width=0.45\textwidth]{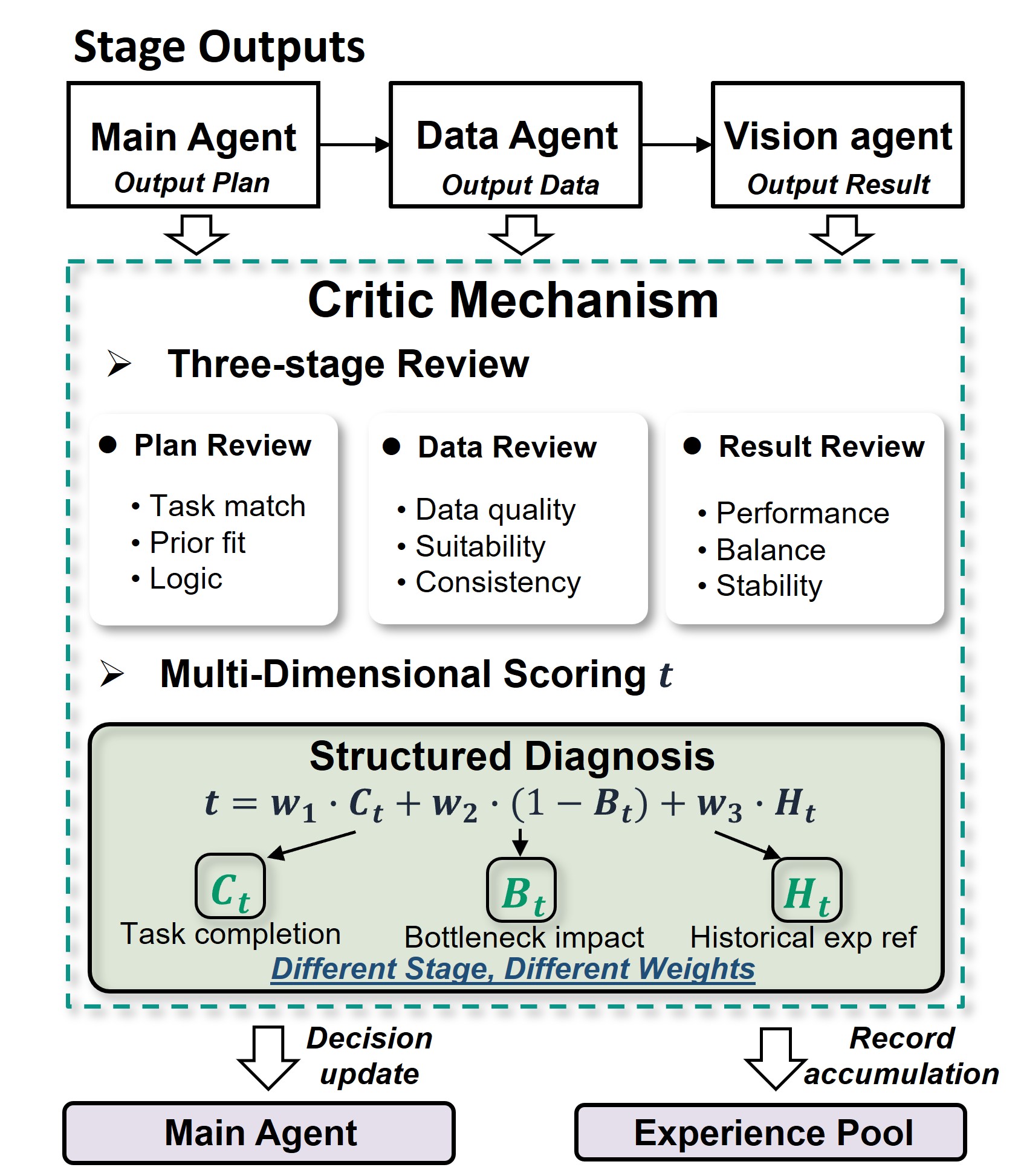}
    \caption{Illustration of the Critic mechanism. The Critic mechanism intervenes at three key nodes of the execution chain, namely the plan review, the data review, and the result review. All feedback is routed through the Main Agent for global coordination, and every iteration is written into the global experience pool after standardization.}
    \label{fig:critic_pipeline}
\end{figure}

The Critic mechanism intervenes at three key nodes of the execution chain rather than evaluating only the final result. As illustrated in Fig.~\ref{fig:critic_pipeline}, the three stages focus on different aspects.

\noindent{\textbf{(a) Plan review.}} The Critic mechanism examines the rationality of the technical plan produced by the Main Agent. The review considers: the alignment between the plan and the task requirement, for example whether an appropriate anchor box setting is chosen for an object detection task; the consistency between the plan and the historical records as well as the literature priors in the global experience pool; and the internal logical coherence of the plan, for example whether the loss function matches the data characteristics. The purpose of this stage is to intercept clearly unreasonable plans before execution and to avoid the waste of computation.

\noindent{\textbf{(b) Data review.}} The Critic mechanism examines whether the data processing scheme is appropriate. The review considers: the fit between the preprocessing strategy and the properties of remote sensing data, for example whether the tile size for large images is reasonable and whether class imbalance is sufficiently handled; the integrity and the format compliance of the processed data; and the coordination between the data processing scheme and the technical plan issued by the Main Agent.

\noindent{\textbf{(c) Result review.}} The Critic mechanism performs a comprehensive quantitative evaluation and diagnosis of the training results. The review considers: whether the overall performance reaches the domain baseline recorded in the global experience pool; whether the performance across categories is balanced; whether the training process exhibits overfitting or instability; and whether a substantive improvement is achieved over the previous round.

\subsubsection{Structured Diagnosis and Multi Dimensional Scoring}
\label{sec:critic_score}

To make the review results quantifiable, comparable, and traceable, the Critic mechanism introduces a multi dimensional scoring mechanism that produces a comprehensive evaluation of each reviewed output. The scoring covers three dimensions.

\noindent{\textbf{(a) Task completion.}} $C_t$ reflects the progress of the current stage toward the predefined objective. In plan review, it measures how well the plan covers the task requirement; in data review, it measures the completeness of the data processing; in result review, it measures how closely the performance approaches the target. The value lies in $[0,1]$.

\noindent{\textbf{(b) Bottleneck impact.}} $B_t$ quantifies how strongly the key deficiency of the current stage affects the overall task performance. The core of this dimension is to identify the primary weakness, which may be an unreasonable architecture in the plan stage, an unresolved class imbalance in the data stage, or a severely lagging category in the result stage. A larger $B_t$ indicates a more severe bottleneck. The value lies in $[0,1]$.

\noindent{\textbf{(c) Global experience influence.}}
\textcolor{black}{$H_t$ measures the consistency of the current direction with
the successful and failed experiences retrieved from $\mathcal{M}_{\mathrm{gb}}$
by the retrieval mechanism described in Section~\ref{sec:internal_flow}. These
entries are provided to the Critic Mechanism as reference information, and the
Critic Mechanism assigns $H_t\in[0,1]$, where a higher value indicates stronger
consistency with previously successful directions and greater avoidance of
known failure patterns.}

\textcolor{black}{The comprehensive score is then given by
\begin{equation}
    \mathrm{Score}_t
    = w_1 C_t + w_2 (1-B_t) + w_3 H_t,
    \label{eq:score_t}
\end{equation}}
\textcolor{black}{
where $w_1$, $w_2$, and $w_3$ are stage-dependent weights fixed before
evaluation. Specifically, $(w_1,w_2,w_3)$ is set to
$(0.30,0.50,0.20)$ for plan review, $(0.35,0.40,0.25)$ for data review,
and $(0.50,0.30,0.20)$ for result review. Thus, plan review places greater
emphasis on the severity of the remaining bottleneck, whereas result review
places greater emphasis on task completion. The final score is recomputed by
the controller from the $C_t$, $B_t$, and $H_t$ values returned by the Critic
Agent to ensure consistent scoring.
}

\textcolor{black}{
The review action is determined by explicit rules. If critical evidence is
missing, the output is marked as \texttt{escalate}. Otherwise, it is accepted
when $\mathrm{Score}_t \geq 0.75$ and $B_t \leq 0.30$, returned for revision
when $\mathrm{Score}_t \geq 0.45$ but the acceptance condition is not met, and
rejected otherwise. The Critic Mechanism therefore provides the semantic diagnosis
and component scores, while score aggregation and the final decision follow
fixed rules.
}

\subsubsection{Feedback Driven Decision Update and Global Experience Accumulation}
\label{sec:critic_feedback}

All feedback from the Critic mechanism is routed to the Main Agent, which performs global coordination and decision making, rather than issuing commands directly to subordinate agents. The decision update of the Main Agent after receiving a diagnostic report can be formalized as
\begin{equation}
    o_{\mathrm{main}}^{\,t+1} = \Pi_{\mathrm{update}}\!\left(s_t, \{c_t, \mathrm{Score}_t\}\right),
    \label{eq:decision_update}
\end{equation}
where $s_t$ denotes the system state at the current stage and $\{c_t, \mathrm{Score}_t\}$ denotes the structured diagnosis and the score produced by the Critic mechanism. Meanwhile, the complete record of each iteration, including the technical plan, the experimental configuration, the score, and the diagnostic conclusion, is standardized and written into the global experience pool:
\begin{equation}
    \mathcal{E}_{\mathrm{global}} = \bigcup_{t=1}^{T} \mathrm{Serialize}(c_t).
    \label{eq:global_pool}
\end{equation}
Consequently, every round of feedback from the Critic mechanism drives two processes simultaneously: the immediate optimization of the current task, and the continuous accumulation of the global experience pool. The more tasks the system handles, the richer the pool becomes. In subsequent tasks, the evaluation baseline of the Critic mechanism becomes more accurate, the initial plan produced by the Main Agent is of higher quality, and the overall convergence speed improves steadily.

\begin{figure*}[!t]
    \centering
    \includegraphics[width=0.95\textwidth]{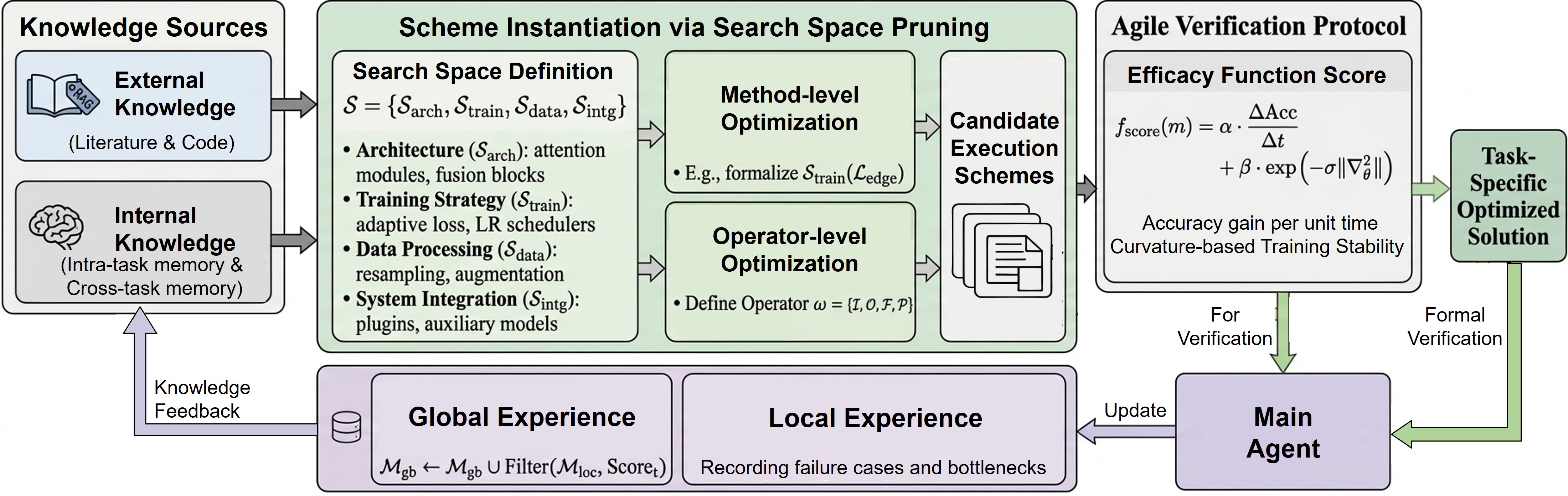}
\caption{
Dual-stream dynamic experience bus for autonomous research evolution. 
External knowledge from literature and code and internal experience from task memory and cross-task memory are integrated into global and local experience pools. 
The experience bus dynamically prunes and expands the structured search space, guides candidate scheme instantiation, and updates the main agent through agile verification, thereby transforming blind trial-and-error search into experience-guided optimization toward task-specific solutions.
}

    \label{fig:evolution}
\end{figure*}
\subsection{Dual-stream Dynamic Experience Bus}
\label{sec:evolution}

To ensure that the system can process diverse visual tasks without falling into blind trial-and-error, we design a dual-stream dynamic experience bus. Through this bus, the system moves from solving a single task to accumulating transferable knowledge across tasks.

\textcolor{black}{As shown in Fig.~\ref{fig:evolution}, we first summarize common optimization
strategies for representative remote sensing tasks and organize them into a
structured search space with four dimensions: model architecture, training
strategy, data processing, and system integration:
\begin{equation}
    \mathcal{S} =
    \{\mathcal{S}_{\mathrm{arch}},\
    \mathcal{S}_{\mathrm{train}},\
    \mathcal{S}_{\mathrm{data}},\
    \mathcal{S}_{\mathrm{intg}}\}.
    \label{eq:search_space}
\end{equation}
Here, $\mathcal{S}_{\mathrm{arch}}$ covers modifications to the backbone,
neck, and feature-fusion modules; $\mathcal{S}_{\mathrm{train}}$ includes
loss functions, optimization strategies, and hyperparameter settings;
$\mathcal{S}_{\mathrm{data}}$ includes resampling, preprocessing, and data
augmentation; and $\mathcal{S}_{\mathrm{intg}}$ covers auxiliary models and
system-level integration strategies. These predefined operators form the base
search space. During evolution, methods retrieved from external literature and
open-source repositories can be incorporated into the corresponding dimension
after filtering and structured parameterization, allowing the search space to
be extended without unconstrained generation by the agents. The detailed
operator definitions and instantiation rules are provided in
Appendix~\ref{app:search_space}.}

\textcolor{black}{The predefined base space contains 23 operators across model architecture,
training strategy, data processing, and system integration. Each operator
specifies its applicable bottleneck tags and a constrained parameter space.
New operators obtained from external retrieval or generated during later
research evolution are not included in this base catalog.}

\subsubsection{External Knowledge Flow}
\label{sec:external_flow}

To address the limitation of fixed knowledge bases in handling complex research logic and emerging research needs, we design an online branch based on retrieval augmented generation. The system retrieves knowledge from external sources, applies differentiated extraction to textual knowledge and code knowledge, and then maps the results jointly into executable instructions in $\mathcal{S}$.

\noindent{\textbf{(a) High level method extraction from text.}} For document type knowledge such as the methodology sections of papers and the README or implementation notes of open source repositories, the system applies semantic distillation with a large language model and focuses on novel loss functions, learning rate schedules, and network connection patterns. These unstructured descriptions are mapped into heuristic guidance vectors within the search space. For instance, if a paper reports that introducing an edge gradient constraint improves detection accuracy, the system formalizes it as an optimization direction $\mathcal{S}_{\mathrm{train}}(\mathcal{L}_{\mathrm{edge}})$ in the training strategy layer.

\noindent{\textbf{(b) Low level operator extraction from code.}} The system does not directly copy source code. It first removes the peripheral code related to input/output, logging, and framework scaffolding, locates the core forward pass, and extracts the essential components into a reusable operator $\omega$. The operator is defined as
\begin{equation}
    \omega = \{\mathcal{I},\ \mathcal{O},\ \mathcal{F},\ \mathcal{P}\},
    \label{eq:operator}
\end{equation}
where $\mathcal{I}$ and $\mathcal{O}$ are the input and output tensors, $\mathcal{F}$ is the core operator logic (for example a BiFPN fusion or a cross attention module), and $\mathcal{P}$ denotes the distribution of key hyperparameters. The agent then analyses the current backbone and automatically searches for candidate mounting positions that match $\omega$.

By combining the evolution directions provided by the textual layer and the standardized operators provided by the code layer, the system generates instantiated execution instructions in $\mathcal{S}$. This mechanism ensures that the latest state of the art paradigms can be integrated on demand.

\subsubsection{Internal Experimental Flow}
\label{sec:internal_flow}

At the end of each iteration, the Critic mechanism conducts an in depth review of the experimental trajectory and uncovers the conditions that lead to accuracy fluctuation or training failure. For the experience knowledge produced in such re experiments, we adopt a hierarchical memory with explicit update logic over different life cycles.

\noindent{\textbf{(a) Local experience within a task.}} This memory integrates the failure cases and the performance bottlenecks reported by the Critic mechanism across the iterations of the current task. Its core role is to construct a negative memory $\mathcal{M}_{\mathrm{loc}}$ that records hyperparameter combinations leading to severe accuracy fluctuation or resource overflow. During the execution of a single task, the memory drives the agent to perform real time path pruning and terminates unpromising attempts as soon as a potential error direction is detected, for example when a training round fails to reach the historical level.

\noindent{\textbf{(b) Global experience across tasks.}} This memory serves as the cognitive center of the system. It organizes experimental data into transferable research knowledge and is never cleared after a task is completed. Based on the stagewise completion score produced by the Critic mechanism, the local negative memory is weighted to form a general prior for the experimental phenomena of the task, which is stored as $\mathcal{M}_{\mathrm{gb}}$. The incremental update is expressed as
\begin{equation}
    \mathcal{M}_{\mathrm{gb}} \leftarrow \mathcal{M}_{\mathrm{gb}} \cup \mathrm{Filter}(\mathcal{M}_{\mathrm{loc}},\ \mathrm{Score}_t),
    \label{eq:mem_update}
\end{equation}
where $\mathrm{Score}_t$ is the comprehensive score from Section~\ref{sec:critic_score} and serves as the quality threshold for admission into the global memory, and $\mathrm{Filter}$ denotes the score based noise filtering that guarantees that only high quality or instructive paths are retained. \textcolor{black}{When facing a new task, the system retrieves relevant successful experiences and failure records from $\mathcal{M}_{\mathrm{gb}}$ according to a weighted task-similarity measure that considers the task description, current bottleneck, dataset characteristics, model, and primary evaluation metric. The detailed similarity computation and retrieval thresholds are provided in Appendix~\ref{app:task_similarity}.}

\subsubsection{Decision Evolution under Experience Guidance and the Agile Verification Protocol}
\label{sec:agile_protocol}

\textcolor{black}{Guided by the dual-stream experience bus, each candidate strategy is first evaluated by a constrained five-minute training run:
\begin{equation}
    f_{\mathrm{score}}(m)
    =
    \alpha \frac{\Delta \mathrm{Acc}}{\Delta t}
    +
    \beta \exp\left(-\sigma \lVert \nabla_{\theta}^{2} \rVert\right),
    \label{eq:f_score}
\end{equation}
where $\Delta \mathrm{Acc}$ denotes the task-specific performance gain and the second term characterizes training stability. In practice, the coefficient of variation of the final 50 training losses is used as a lightweight surrogate for the stability term. $\Delta \mathrm{Acc}$ is expressed in percentage points, $\Delta t$ is normalized to one verification window.}

\textcolor{black}{A candidate is promoted to full training only if it achieves a positive performance gain, completes verification without training failure, satisfies $f_{\mathrm{score}}(m)>0.55$, passes the result-stage Critic check, and is not rejected by negative memory. Otherwise, the candidate is discarded and the evolution continues. The evolution terminates when the maximum of 20 rounds is reached or no candidate improves the best-so-far result for two consecutive full-experiment rounds.}

\section{Experiments}
\begin{table*}[t]
\centering
\caption{\textcolor{black}{Object detection results on FAIR1M (\%). Overall mAP$_{50}$ and per-category AP are reported for selected weak categories. Representative detectors are compared with the baseline and successive evolution stages of RingMoClaw. Best results are in \textbf{bold}.}}
\label{tab:fair1m_weak_only}
\resizebox{\textwidth}{!}{
\begin{tabular}{l|c|ccccccccc}
\midrule
\textbf{Method} 
& \textbf{mAP$_{50}$}
& \textbf{C5}
& \textbf{C10}
& \textbf{C11}
& \textbf{C14}
& \textbf{C24}
& \textbf{C25}
& \textbf{C26}
& \textbf{C27}
& \textbf{C33} \\
\midrule

Gliding Vertex~\cite{xu2021glidingvertex}
& 29.92 & 0.01 & 0.01 & 9.12 & 15.67 & 14.32 & \textbf{16.39} & 16.92 & 28.91 & 16.25 \\

RetinaNet~\cite{lin2017focal}
& 30.67 & 0.81 & 1.70 & 9.57 & 16.37 & 19.17 & 1.28 & 17.03 & 28.98 & 17.41 \\

Cascade R-CNN~\cite{cai2018cascade}
& 31.18 & 0.00 & 0.00 & 12.10 & 15.35 & 13.66 & 0.91 & 16.45 & 30.27 & 20.35 \\

Faster R-CNN~\cite{ren2015faster}
& 32.12 & 0.01 & 0.13 & 9.81 & 17.65 & 15.03 & 3.04 & 17.99 & 29.36 & 16.92 \\

RoI Transformer~\cite{ding2019learning}
& 35.29 & 0.00 & 0.00 & 14.31 & 14.32 & 16.22 & 5.13 & 22.17 & 46.71 & 18.58 \\

OFA-Net~\cite{hu2021ofa}
& 43.73 & 30.71 & 29.25 & 9.90 & 30.37 & 13.45 & 6.06 & 17.36 & 5.35 & 24.68 \\

Oriented R-CNN~\cite{xie2021oriented}
& 44.30 & 25.55 & 32.85 & 20.46 & \textbf{32.89} & 14.50 & 4.42 & 19.81 & 7.51 & 21.87 \\

\midrule
\multicolumn{11}{l}{\textbf{RingMoClaw evolution}} \\
\midrule

Baseline (Train \& Infer)
& 47.30 & 51.74 & 32.56 & 31.11 & 20.29 & 26.55 & 3.29 & 37.42 & 52.81 & 42.41 \\

Data Augmentation
& 48.18 & 51.94 & 34.76 & 31.14 & 21.89 & 33.75 & 3.29 & 41.38 & 52.81 & 45.11 \\

Training Config Update
& 47.77 & 51.83 & 35.06 & 31.13 & 23.90 & 34.95 & 3.42 & 41.42 & \textbf{52.92} & 45.16 \\

Few-Shot Training
& 47.69 & 51.92 & 35.15 & 31.58 & 24.11 & 32.89 & 3.74 & 41.66 & 51.77 & 45.20 \\

Multi-Model Fusion
& 49.11 & 53.87 & 36.27 & 32.87 & 26.60 & 34.20 & 4.01 & 41.64 & 52.15 & \textbf{45.54} \\

\textbf{Module Innovation}
& \textbf{49.14} & \textbf{54.14} & \textbf{36.36} & \textbf{33.16}
& 26.89 & \textbf{35.33} & 4.74 & \textbf{42.03}
& 52.14 & 45.29 \\

\midrule
\end{tabular}
}

\vspace{1mm}
\begin{minipage}{\textwidth}
\footnotesize
\textit{Note:} C5: C919; C10: ARJ21; C11: Passenger Ship; C14: Tugboat;
C24: Trailer; C25: Tractor; C26: Excavator; C27: Truck Tractor;
C33: Roundabout.
\end{minipage}

\end{table*}

\begin{figure*}[t]
\centering
\includegraphics[width=1.0\linewidth]{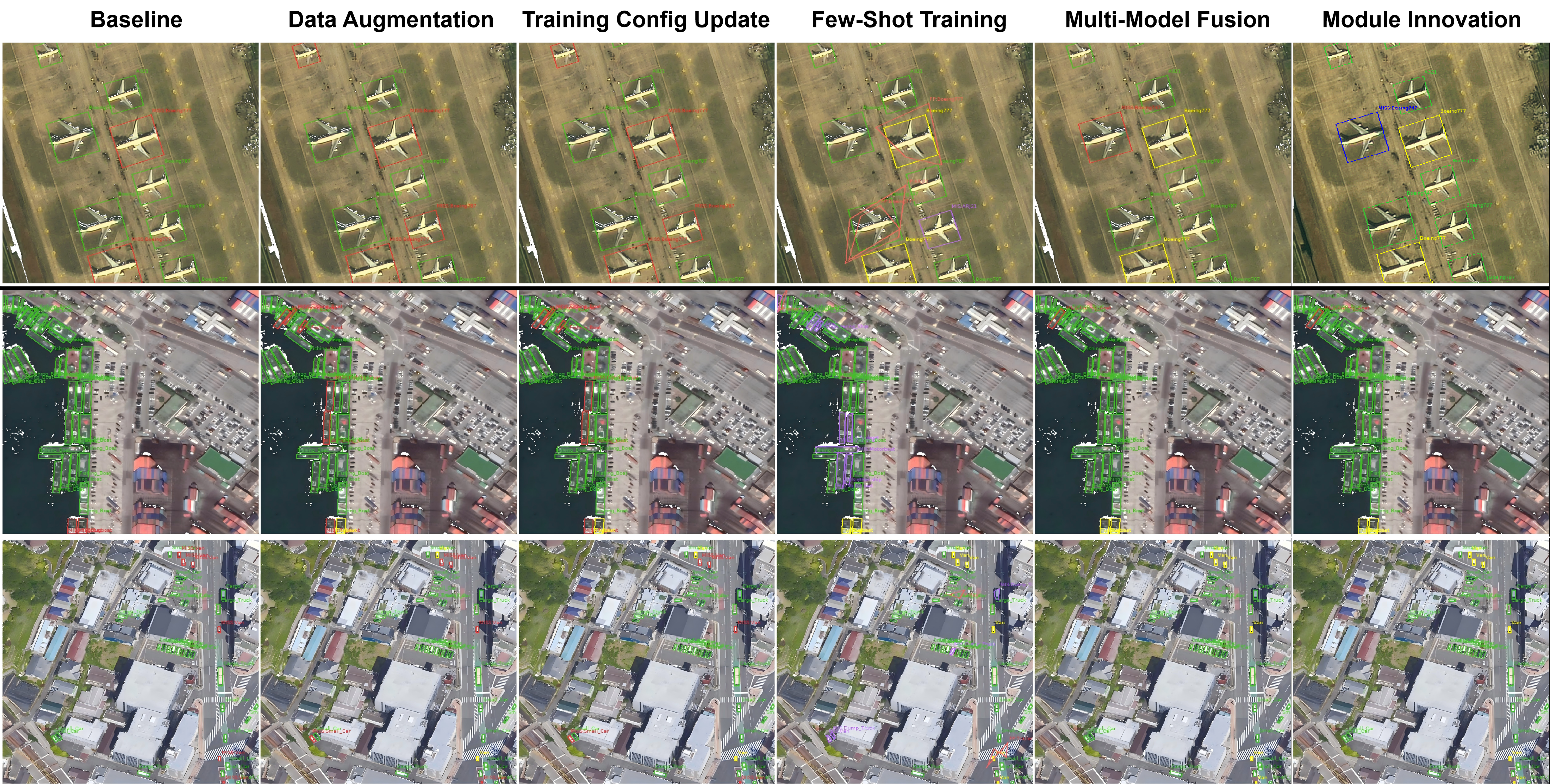}
\caption{Weak categories detection results on FAIR1M across different evolution stages. From left to right: (1) Baseline, (2) Data Augmentation, (3) Training Strategy Optimization, (4) Two-Stage Few-Shot Detection, (5) Multi-Model Fusion, and (6) Module Innovation. Green boxes denote true positives (TP) of base categories; yellow/purple boxes denote true positives (TP) of weak/rare categories; red boxes denote false negatives (FN, missed detections); and pink boxes denote false positives (FP, incorrect detections). The figure illustrates how progressive improvements in architecture and module design reduce both missed detections and false positives, particularly for weak categories, highlighting the effectiveness of the proposed RingMoClaw Auto-Research framework in addressing long-tail fine-grained detection challenges.}
\label{fig:qualitative_compare}
\end{figure*}

\begin{figure}[t]
\centering
\includegraphics[width=0.7\linewidth]{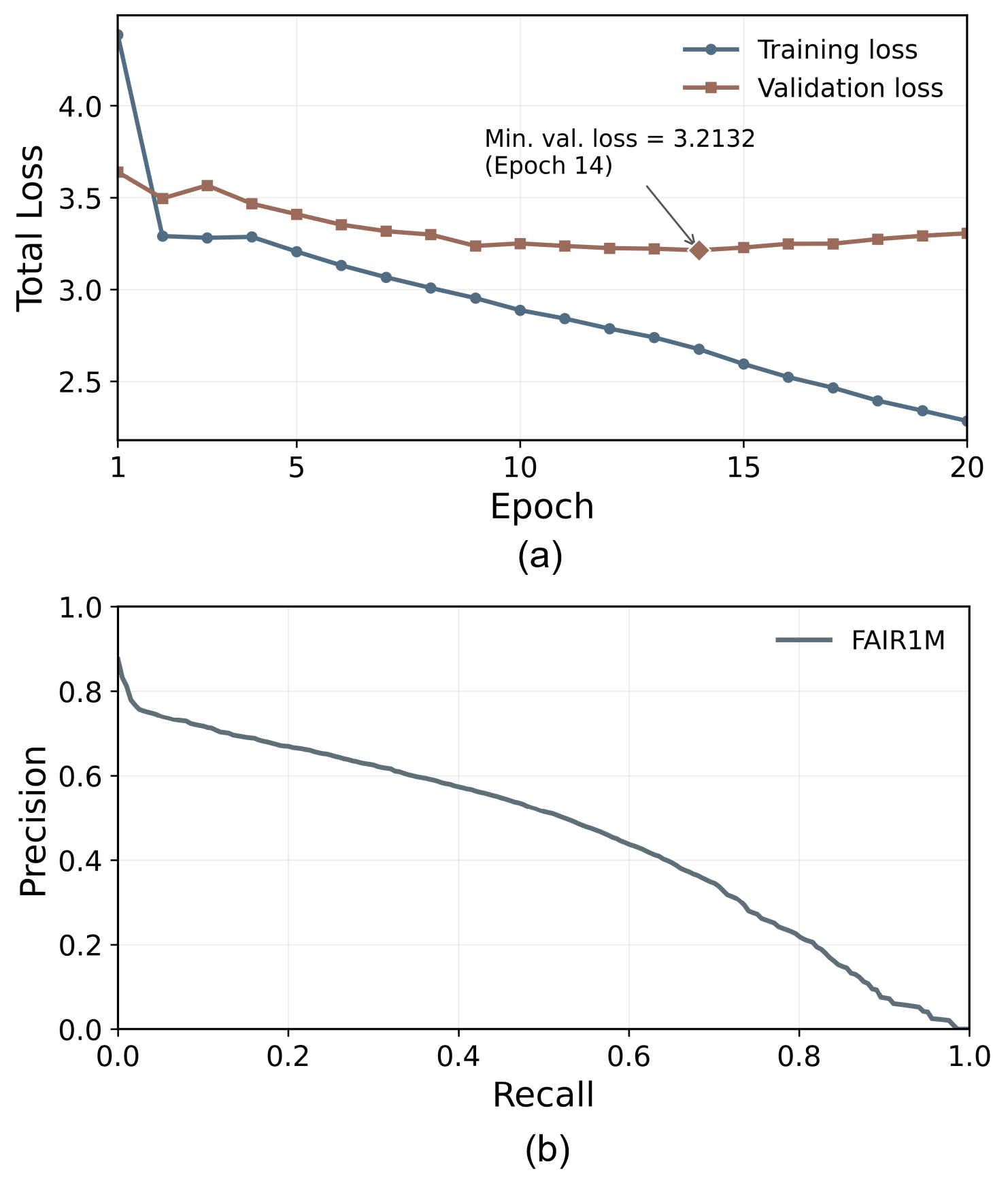}
\caption{\textcolor{black}{
Training behavior of the final detector on FAIR1M.
(a) Training and validation loss curves over 20 epochs.
(b) Mean precision-recall curve over the 37 object categories.
}}
\label{fig:loss_pr}
\end{figure}

In this section, we evaluate the proposed framework from four complementary
perspectives. We first describe the experimental settings, including the
datasets, baselines, and evaluation protocols (Section~\ref{sec:exp_settings}).
We then report the overall performance and the evolution process on four
remote sensing visual tasks, together with a cross-task comparison against
existing research automation frameworks (Section~\ref{sec:exp_compare}), and
analyze the practical cost and search efficiency
(Section~\ref{sec:cost}). Finally, we conduct ablation studies on the Critic
mechanism and the dual-stream experience bus, and validate the five-minute
verification protocol (Section~\ref{sec:ablation}).

\subsection{Experimental Settings}
\label{sec:exp_settings}



\textcolor{black}{We evaluate the proposed framework on four representative remote sensing
visual tasks: oriented object detection, scene
classification, semantic segmentation, and change detection. The datasets and
the task specific settings are summarized below.
RingMo~\cite{sun2022ringmo} is adopted as the primary backbone,
leveraging its robust representations obtained via masked image modeling
pre-training. All experiments are conducted on two NVIDIA RTX 4090 GPUs.}

The agents in the framework are driven by large language models through
their official APIs. The Main Agent, Data Agent, and Vision Agent in the
research branch are all built on GLM-5 Turbo~\cite{zeng2026glm}. The Critic mechanism is
independently powered by Minimax 2.7~\cite{chen2026minimax}. The two models differ in
pretraining data and reasoning style, which keeps the review branch
heterogeneous with respect to the execution chain.

\textcolor{black}{\textbf{Object detection.} We report mean average precision ($mAP_{50}$) as the evaluation metric. The baseline detector adopts a YOLO11x-OBB oriented detection head, following the FAIR1M detection setting in~\cite{sun2022ringmo}. On FAIR1M-v1.0, high-resolution images are
cropped into $512 \times 512$ patches with a 128-pixel overlap. Training runs for 20 epochs with an initial learning rate of
$1 \times 10^{-2}$, following a standard training schedule.}


\textcolor{black}{\textbf{Scene classification.}} 
\textcolor{black}{We report Top-1 accuracy (Acc) as the evaluation metric. NWPU-RESISC45~\cite{cheng2017remote} contains 45 scene categories with $256 \times 256$ images. We split the dataset into 6,300 training, 12,600 validation, and 12,600 test images, following a 1:2:2 ratio. During training, images are randomly cropped to $224 \times 224$ and augmented with horizontal flipping and color jittering, followed by ImageNet normalization. Each classifier is trained for 50 epochs with an initial learning rate of $1 \times 10^{-3}$ under a cosine annealing schedule.}

\textcolor{black}{\textbf{Change Detection.}}
\textcolor{black}{We report the F1 score and IoU of the changed class as the evaluation metrics. LEVIR-CD~\cite{stanet} contains 637 pairs of bi-temporal remote sensing images with a spatial resolution of $1024 \times 1024$, which are split into 445, 64, and 128 pairs for training, validation, and testing, respectively. Following the standard protocol, each image pair is cropped into $256 \times 256$ patches for training. Random horizontal flipping, vertical flipping, and temporal swapping are applied for data augmentation, and the input images are normalized using ImageNet statistics. All models are trained under an identical schedule and selected according to the best validation F1 score.}

\textcolor{black}{\textbf{Semantic Segmentation.}}
\textcolor{black}{We use mean Intersection over Union (mIoU) as the evaluation metric. Experiments are conducted on iSAID~\cite{zamir2019isaid}, with images cropped into $896\times896$ patches with an overlap of 384 pixels. Training employs random resizing, cropping, flipping, and photometric distortion for data augmentation. We use AdamW with an initial learning rate of $6\times10^{-5}$ and a weight decay of 0.01, together with a 1,500-iteration linear warm-up followed by polynomial decay. All models are trained for 80k iterations with an effective batch size of 16 using mixed precision.}

\subsection{Overall Performance and Evolution Process}
\label{sec:exp_compare}

\subsubsection{Object Detection}
\textcolor{black}{We use object detection as a representative task to examine whether RingMoClaw can adapt its optimization strategy to task-specific bottlenecks. Experiments are conducted on FAIR1M, which exhibits a long-tailed category distribution and high inter-class similarity. The evolution process was independently repeated three times, and Table~\ref{tab:fair1m_weak_only} reports the trajectory of the run with the highest final mAP$_{50}$.}

\textcolor{black}{As shown in Table~\ref{tab:fair1m_weak_only}, the overall mAP$_{50}$ improves from 47.30\% to 49.14\%. Data augmentation and training configuration adjustment bring moderate gains, while few-shot training yields no further improvement and prompts the system to move toward model-level optimization. Multi-model fusion raises mAP$_{50}$ to 49.11\%. The module innovation stage evaluates the two generated modules separately and in combination through three trials, and the combined configuration reaches 49.14\%. Detailed results are reported in Sec.~\ref{sec:module innovation}. Since this stage is summarized in a single row, the complete trajectory contains seven evolution trials. Fig.~\ref{fig:qualitative_compare} shows the detection results of weak categories along the same trajectory, where missed detections and false positives decrease as the evolution proceeds. The consistent gains on several weak categories indicate that RingMoClaw adjusts its optimization path according to intermediate experimental feedback.}


\textcolor{black}{
We further examine the convergence and potential overfitting behavior
of the detector, as shown in Fig.~\ref{fig:loss_pr}. The training loss
decreases consistently throughout the 20 epoch schedule, while the
validation loss decreases from 3.6387 to a minimum of 3.2132 at epoch
14 and then slightly increases to 3.3053 at epoch 20. This indicates
only a mild late-stage overfitting tendency. Since only six epochs
remain after the minimum validation loss is reached, patience values
of 10 or 20 would not trigger before completion of the prescribed
training schedule and therefore provide little additional
computational saving. The mean precision-recall curve over the 37
FAIR1M categories is also provided in Fig.~\ref{fig:loss_pr}(b).
}

\begin{table*}[ht]
\centering

\caption{\textcolor{black}{Scene classification results on NWPU-RESISC45 (\%). The table reports overall accuracy and per-category accuracy for selected weak categories. Representative classifiers are compared with the baseline and fusion strategies selected by RingMoClaw. Best results are in \textbf{bold}.}}
\label{tab:nwpu_resisc45_results}
\resizebox{\textwidth}{!}{%
\begin{tabular}{l|c|ccccccccc}
\toprule
Method & Acc & C7 & C10 & C14 & C18 & C23 & C27 & C30 & C34 & C43 \\
\midrule
ConvNeXt-T \cite{liu2022convnet}& 92.25 & 80.71 & 84.29 & 86.07 & 88.21 & 85.36 & 67.50 & \textbf{86.07} & 90.36 & 87.86 \\
EfficientNet-B4 \cite{tan2019efficientnet} & 92.08 & 77.86 & 93.93 & 87.86 & 80.36 & 86.43 & 65.71 & 80.00 & 93.21 & 91.43 \\
ResNet50 \cite{he2016deep} & 82.10 & 46.07 & 77.50 & 70.36 & 77.50 & 56.43 & 58.57 & 61.79 & 86.43 & 71.79 \\
\midrule
\multicolumn{11}{l}{\textbf{RingMoClaw Evolution}} \\
\midrule
Baseline (Train \& Infer)& 93.18 & 73.93 & 91.07 & 87.14 & 88.93 & 77.86 & 75.00 & 83.57 & 90.36 & 87.86 \\
Multi-Model Fusion & 93.71 & 80.71 & 93.93 & 87.14 & 88.21 & 85.36 & 75.00 & \textbf{86.07} & 93.21 & 91.43 \\
Soft Multi-Model Fusion & 94.48 & 80.36 & \textbf{95.00} & 89.29 & 91.07 & 86.43 & \textbf{75.36} & 85.36 & 93.93 & 90.00 \\
\textbf{Soft TTA Fusion} & \textbf{94.97} & \textbf{82.29} & \textbf{95.00} & \textbf{92.14} & \textbf{92.50} & \textbf{90.36} & 72.86 & 84.29 & \textbf{94.64} & \textbf{92.86} \\
\bottomrule
\end{tabular}%
}

\begin{flushleft}
\noindent\footnotesize{*The correspondence between weak categories: 
C7: church, C10: commercial\_area, C14: freeway, C18: industrial\_area, 
C23: medium\_residential, C27: palace, C30: railway\_station, 
C34: runway, C43: thermal\_power\_station.}
\end{flushleft}
\end{table*}

\begin{table}[t]
\centering
\caption{\textcolor{black}{Comparison with representative remote sensing semantic segmentation methods and the evolution results of RingMoClaw on iSAID validation set. Best results are in \textbf{bold}.}}
\label{tab:isaid_segmentation}
\small
\renewcommand{\arraystretch}{1.05}

\begin{tabular*}{\columnwidth}
{@{\extracolsep{\fill}}l|ccc|c@{}}
\toprule
Method / Strategy & LV & RA & SV & mIoU (\%) \\
\midrule
RANet~\cite{shen2020ranet}
    & 60.1 & 70.5 & 49.3 & 62.1 \\
HMANet~\cite{niu2021hybrid}
    & 59.7 & 62.9 & 50.3 & 62.6 \\
FarSeg~\cite{zheng2020foreground}
    & 60.6 & 71.4 & 46.3 & 63.7 \\
FactSeg~\cite{ma2021factseg}
    & 62.7 & 69.4 & 49.5 & 64.8 \\
GRDL~\cite{niu2021improving}
    & 60.5 & 68.7 & 50.7 & 65.1 \\
UperNet (RSP-ViTAEv2-S)~\cite{wang2022empirical}
    & 62.5 & 62.2 & 51.6 & 64.3 \\
UperNet-RingMo~\cite{sun2022ringmo}
    & 63.9 & 67.3 & 51.2 & 67.2 \\
\midrule
\multicolumn{5}{l}{\textbf{RingMoClaw Evolution}} \\
\midrule
Baseline (Train \& Infer)
    & 64.8 & 64.9 & 50.4 & 67.2 \\
Larger Backbone
    & 65.5 & \textbf{72.9} & 51.8 & 69.0 \\
Single-scale Upscaling
    & \textbf{67.7} & 69.2 & \textbf{54.9} & 69.5 \\
Multi-scale TTA
    & 67.0 & 72.4 & 53.7 & 69.8 \\
\textbf{Upscale-only TTA}
    & 67.7 & 70.6 & 54.8 & \textbf{69.9} \\
\bottomrule
\end{tabular*}

\begin{flushleft}
\noindent\footnotesize{*LV, RA, and SV denote Large Vehicle, Roundabout, and Small Vehicle, respectively.}
\end{flushleft}

\end{table}

\subsubsection{Scene Classification}

\begin{figure*}[t]
\centering
\includegraphics[width=1.0\linewidth]{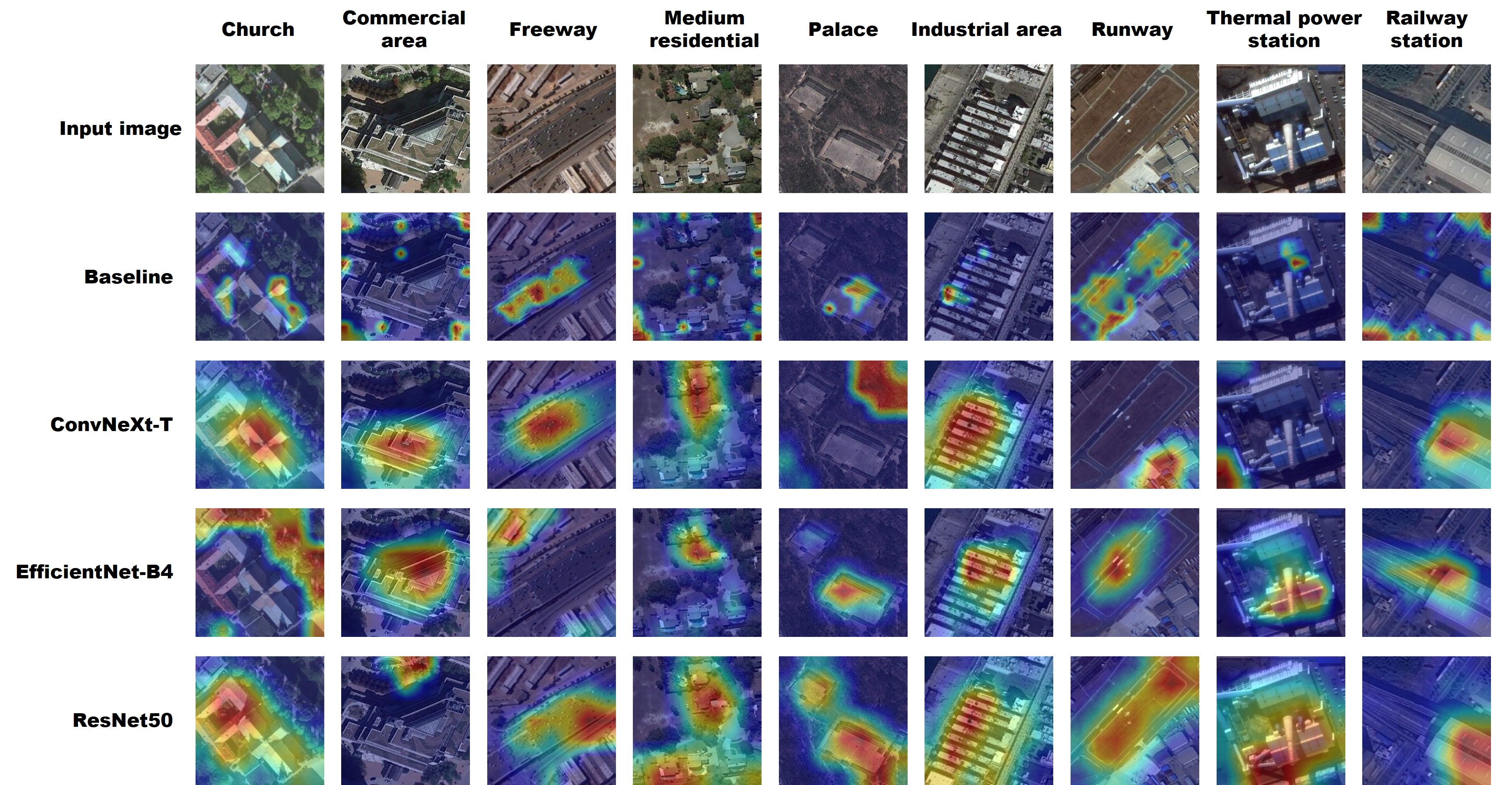}
\caption{Class activation map visualization of the baseline classifier and the candidate classifiers selected for score-level fusion on representative weak categories of NWPU-RESISC45. Different candidate models attend to different discriminative regions, indicating clear complementary behavior across scene categories.}
\label{fig:nwpu_cam}
\end{figure*}


\textcolor{black}{Table~\ref{tab:nwpu_resisc45_results} reports the classification results on NWPU-RESISC45. The baseline achieves an overall accuracy of 93.18\%, while the selected candidate classifiers, ConvNeXt-T, EfficientNet-B4, and ResNet50, obtain 92.25\%, 92.08\%, and 82.10\%, respectively. Although no individual candidate consistently surpasses the baseline, their per-category results reveal complementary strengths across different scene types. Based on this observation, decision-level fusion is adopted to integrate their predictions. Multi-Model Fusion continuously improves the accuracy from 93.71\% to 94.48\%, and Soft Multi-Model Fusion further achieves the best overall accuracy of 94.97\%, outperforming the baseline by 1.79 percentage points. These results demonstrate that score-level soft fusion effectively exploits the complementary predictions of different classifiers for scene classification.}

\textcolor{black}{Fig.~\ref{fig:nwpu_cam} visualizes the class activation maps of the baseline model and three candidate classifiers on representative weak categories of NWPU-RESISC45. The results show that different models focus on different discriminative regions even for the same scene category. For example, some models emphasize dominant objects such as runways or industrial structures, while others attend more to surrounding contextual regions. This indicates that the candidate classifiers provide complementary visual evidence rather than redundant responses.}

\subsubsection{\textcolor{black}{Semantic Segmentation}}
\textcolor{black}{As shown in Table~\ref{tab:isaid_segmentation}, RingMoClaw progressively improves the segmentation performance on iSAID from 67.2\% to 69.9\% mIoU. A larger backbone first increases the mIoU to 69.0\%, while subsequent scale-aware inference further improves the overall performance. In particular, single-scale upscaling improves the IoU of LV and SV to 67.7\% and 54.9\%, respectively, and the final upscale-only TTA achieves the best overall result of 69.9\% mIoU.}

\subsubsection{\textcolor{black}{Change Detection}}
\textcolor{black}{As shown in Table~\ref{tab:change_detection}, RingMoClaw improves the change detection performance on the LEVIR-CD validation set from 89.53\% F1 and 87.10\% IoU to 90.54\% F1 and 89.97\% IoU. Temporal-Swap Augmentation first improves the IoU to 87.97\%, indicating enhanced robustness to the temporal ordering of image pairs. Bidirectional ChangeStar then raises the F1 and IoU to 90.18\% and 88.89\%, respectively, by explicitly modeling bidirectional temporal interactions. Although Deep Change Interaction alone yields 89.65\% F1 and 88.57\% IoU, the final EvoChangeStar achieves the best overall performance, reaching 90.54\% F1 and 89.97\% IoU.}

\begin{table}[t]
\centering
\caption{\textcolor{black}{Comparison with representative remote sensing change detection methods and the evolution results of RingMoClaw on LEVIR-CD validation set. Best results are in \textbf{bold}.}}
\label{tab:change_detection}
\small
\renewcommand{\arraystretch}{1.1}

\begin{tabular}{l|cc}
\toprule
\bf Method / Strategy & \bf F1  & \bf IoU \\
\midrule
FC-Siam-Di~\cite{fcdi}  & 86.31 & 83.31 \\
FC-Siam-Conc~\cite{fcdi}  & 83.69 & 76.77 \\
IFNet~\cite{ZHANG2020183}  & 88.13 & 82.93 \\
STANet~\cite{stanet}     & 87.30 & 77.40 \\
BIT~\cite{bit}           & 89.31 & 89.37 \\
SNUNet~\cite{snunet}        & 88.16 & 87.17 \\
ChangeFormer~\cite{changeformer}  & 90.40 & 88.80 \\
\midrule
\multicolumn{3}{l}{\textbf{RingMoClaw Evolution}} \\
\midrule
Baseline (Train \& Infer)        & 89.53 & 87.10 \\
Temporal-Swap Augmentation  & 88.91 & 87.97 \\
Bidirectional ChangeStar  & 90.18 & 88.89 \\
Deep Change Interaction  & 89.65 & 88.57 \\
\textbf{EvoChangeStar} & \textbf{90.54} & \textbf{89.97} \\
\bottomrule
\end{tabular}
\end{table}

\subsubsection{\textcolor{black}{Comparison with Automated Research Frameworks}}
\begin{table*}[t]
\centering
\caption{\textcolor{black}{Cross-task comparison with representative research-automation frameworks. Performance gains are computed relative to the corresponding baseline model for each task.}}
\label{tab:framework_comparison}
\resizebox{\textwidth}{!}{
\begin{tabular}{lccc|cc|cc|ccc}
\midrule
\multirow{2}{*}{\textbf{Framework}}
& \multicolumn{3}{c|}{\textbf{Object Detection}}
& \multicolumn{2}{c|}{\textbf{Scene Classification}}
& \multicolumn{2}{c|}{\textbf{Semantic Segmentation}}
& \multicolumn{3}{c}{\textbf{Change Detection}} \\
\cline{2-11}
\noalign{\vskip 2pt}
& $\Delta$mAP$_{50}$ $\uparrow$
& $\Delta$mAP$_{50:95}$ $\uparrow$
& ER $\downarrow$
& $\Delta$Acc $\uparrow$
& ER $\downarrow$
& $\Delta$mIoU $\uparrow$
& ER $\downarrow$
& $\Delta$F1 $\uparrow$
& $\Delta$IoU $\uparrow$
& ER $\downarrow$ \\
\midrule

AutoResearch~\cite{karpathy2026autoresearch}
& -3.50
& -1.13
& 20
& 1.00
& 11
& 1.02
& 8
& -0.98
& 0.16
& 14 \\

AutoResearchClaw~\cite{liu2026autoresearchclaw}
& 0.57
& 0.12
& 18
& 1.31
& 10
& 1.48
& 7
& 0.24
& 1.08
& 12 \\

\textbf{RingMoClaw}
& \textbf{1.84}
& \textbf{0.51}
& \textbf{7}
& \textbf{1.79}
& \textbf{4}
& \textbf{2.70}
& \textbf{4}
& \textbf{1.01}
& \textbf{2.87}
& \textbf{6} \\

\midrule
\end{tabular}
}

\vspace{1mm}
\begin{minipage}{\textwidth}
\footnotesize
\textit{Note:}
Performance gain is defined as the difference between the final evolved model and the corresponding baseline model, i.e.,
$\Delta P = P_{\mathrm{final}} - P_{\mathrm{baseline}}$.
``ER'' denotes the number of completed evolution steps required by each framework on the corresponding task.
Higher performance gains and fewer evolution rounds indicate better research-automation effectiveness and efficiency.
\end{minipage}

\end{table*}

\textcolor{black}{To evaluate RingMoClaw at the research-automation level, we further compare it with AutoResearch~\cite{karpathy2026autoresearch} and AutoResearchClaw~\cite{liu2026autoresearchclaw} across four representative remote sensing tasks. As shown in Table~\ref{tab:framework_comparison}, AutoResearch adopts a single-agent linear keep/discard loop based on short-budget validation, while AutoResearchClaw organizes research into a staged forward pipeline. For a fair comparison focused on model optimization, we retain AutoResearchClaw only through the experiment-execution stage and exclude its subsequent paper generation and publishing steps. In contrast, RingMoClaw forms a closed-loop evolution process through experience retrieval, heterogeneous Critic review, agile verification, and memory update. The results show that this feedback-driven design enables more effective optimization with fewer evolution rounds, indicating improved research efficiency rather than simply increased experimental exploration.}

\subsubsection{Progressive Analysis of the Evolution Process}

\paragraph{Data Augmentation}
The first stage focuses on data-level intervention. After the initial task analysis, the Data Agent identifies the scarcity of several fine-grained categories and constructs an augmented training set for these categories. Conventional transformations, including rotation, flipping, brightness perturbation, Gaussian noise injection, and affine transformation, are applied to increase the visual diversity of rare-class samples. The Vision Agent then retrains the detector using the augmented data and reports the updated detection results to the Critic mechanism. The Critic mechanism evaluates whether the augmented samples lead to meaningful improvement and records the limited effect of this direction in the experience pool.

\paragraph{Training Strategy Optimization}
After the data-level strategy is evaluated, the framework proceeds to training-level optimization while keeping the model architecture unchanged. The Main Agent updates the training configuration according to the feedback from the previous stage. The adopted adjustments include class-aware sampling, category reweighting, focal loss, and schedule refinement, which aim to increase the exposure of underrepresented categories and emphasize hard samples during optimization. The Vision Agent executes the updated training plan, and the Critic mechanism reviews the resulting performance and category-level behavior. This stage allows the system to examine whether the weak-category bottleneck can be alleviated through optimization-level adjustment.

\paragraph{Two-Stage Few-Shot Detection}
When data augmentation and training configuration updates provide limited improvement, the framework explores a task-decomposition strategy. The Main Agent reformulates the original fine-grained detection task into a two-stage pipeline. In the first stage, aircraft subcategories are merged into a unified airplane category for object localization. In the second stage, the detected aircraft regions are cropped and sent to a few-shot classifier for fine-grained recognition. This design attempts to separate localization from fine-grained category discrimination. The Critic mechanism then analyzes the output of this pipeline and identifies that the final performance is constrained by the recall of the first-stage detector, because missed rare targets cannot be recovered by the subsequent classifier.

\paragraph{Multi-Model Fusion}
Based on the limitations observed in the previous stages, RingMoClaw further explores model-level complementarity. The Main Agent selects complementary detectors from the candidate model pool, and the Vision Agent evaluates their predictions under the same task setting. A category-aware fusion strategy is then constructed to combine their outputs. This stage is designed to test whether different detectors can compensate for each other on categories with different visual patterns. The Critic mechanism evaluates the fused predictions and records both the benefit of model complementarity and the additional deployment cost introduced by multi-model inference.

\paragraph{Module Innovation}
\label{sec:module innovation}
\textcolor{black}{Finally, the framework extends the evolution process from selecting and combining existing strategies to architecture-level exploration.
According to the bottlenecks identified in previous rounds, the Main Agent generated two task-specific architectural hypotheses, termed Multi-Scale Feature Refinement (MSFR) and Contrastive Prototype Refinement (CPR), and instantiated them on Oriented R-CNN. Specifically, MSFR replaces the standard FPN neck with an interface-compatible multi-scale refinement module, whereas CPR augments the rotated RoI classification head with prototype-based logit refinement. The Vision Agent implements and evaluates these candidates, while the Critic Mechanism determines whether the resulting architectural modifications should be retained. Table~\ref{tab:module_innovation} reports the performance and computational overhead of the individual and combined module configurations. Detailed architectures, mathematical formulations, insertion positions, and implementation details are provided in Appendix~\ref{app:module_details}.}

\begin{table}[t]
\centering
\caption{\textcolor{black}{Evaluation of the MSFR and CPR.}}
\label{tab:module_innovation}
\resizebox{\columnwidth}{!}{
\begin{tabular}{lcccccc}
\toprule
\textbf{Configuration} &
\textbf{MSFR} &
\textbf{CPR} &
\textbf{mAP$_{50}$} &
\textbf{mAP$_{50:95}$} &
\textbf{Params} &
\textbf{FLOPs} \\
\midrule
Original baseline
& -- & -- & 47.30 & 36.59 & 119 M & 167 G \\

\midrule
MSFR
& \checkmark & -- & 48.31 & 36.77 & +0.599 M & +0.604 G \\

CPR
& -- & \checkmark & 48.29 & 36.72 & +0.131 M & +0.272 G \\

MSFR + CPR
& \checkmark & \checkmark & \textbf{49.14} & \textbf{37.10}
& +0.730 M & +0.876 G \\
\bottomrule
\end{tabular}}
\end{table}

\textcolor{black}{The results show that the jointly retained MSFR-CPR configuration achieves mAP$_{50}$/mAP$_{50:95}$ 49.14/37.10 while introducing only 0.730~M additional parameters and 0.876~G FLOPs.}

\subsection{Practical Cost and Search Efficiency}
\label{sec:cost}



\begin{table*}[t]
\centering
\caption{\textcolor{black}{Practical cost and search efficiency comparison of autonomous research frameworks on the object detection task.}}
\label{tab:cost_comparison}
\small
\resizebox{\textwidth}{!}{
\begin{tabular}{lcccccc}
\toprule
Framework &
Evolution Rounds &
Explored Candidates &
GPU-hours &
Wall-clock (h) &
Input Tokens (M) &
Output Tokens (M)\\
\midrule
AutoResearch
& 20 & 20 & 26.90 & 13.45 & 1.16 & 0.05 \\
AutoResearchClaw
& 18 & 35 & 18.28 & 13.28 & 2.68 & 0.11\\
RingMoClaw
& 7 & 28 & 12.14 & 10.61 & 0.72 & 0.38 \\
\bottomrule
\end{tabular}
}

\par\vspace{2pt}
\footnotesize
``Explored Candidates'' denotes the total number of optimization proposals considered during evolution. AutoResearch generates one candidate per round, whereas AutoResearchClaw and RingMoClaw may consider multiple candidates within a round. All LLM calls were conducted through CSTCloud and incurred no commercial API cost.
\end{table*}



\textcolor{black}{For the object detection task, we further compare RingMoClaw with AutoResearch~\cite{karpathy2026autoresearch} and AutoResearchClaw~\cite{liu2026autoresearchclaw} in terms of evolution rounds, explored candidates, GPU consumption, wall-clock time, and LLM usage. As shown in Table~\ref{tab:cost_comparison}, AutoResearch follows a linear one-candidate-per-round process, whereas AutoResearchClaw and RingMoClaw may consider multiple candidate proposals within each round. Despite the additional reasoning required for planning, Critic-based diagnosis, and experience summarization, RingMoClaw achieves lower overall computational cost by reducing unnecessary experimental trials and accelerating the evolution process.}

\subsection{Ablation Study}
\label{sec:ablation}
\subsubsection{Effectiveness of the Critic mechanism}

\textcolor{black}{To evaluate the contribution of the Critic mechanism, we remove the Critic Mechanism while keeping the remaining framework and experimental settings unchanged. In this variant, the Main Agent determines subsequent evolution rounds without independent diagnostic feedback from the Critic Mechanism.}

\textcolor{black}{As shown in Table~\ref{tab:critic_ablation}, the full framework converges in 7 evolution rounds and achieves a final mAP of 49.14\%, whereas removing the Critic mechanism requires 11 rounds and results in a lower mAP of 47.43\%. These results indicate that the Critic mechanism improves both evolution efficiency and final detection performance by providing structured diagnosis and reducing ineffective exploration.}

\begin{table}[t]
\centering
\caption{\textcolor{black}{Ablation Study on the Critic Mechanism on FAIR1M.}}
\label{tab:critic_ablation}
\begin{tabular}{lcc}
\toprule
Variant & Rounds & Final mAP (\%) \\
\midrule
Full Framework (Ours) & 7 & 49.14 \\
w/o Critic Mechanism & 11 & 47.43 \\
\bottomrule
\end{tabular}
\par\vspace{2pt}
\footnotesize
``Rounds'' denotes the number of full evolution rounds until the detection accuracy no longer shows a clear improvement and the evolution converges. ``Final mAP'' reports the detection accuracy at convergence.
\end{table}

\begin{table}[t]
\centering
\caption{\textcolor{black}{Ablation Study on the Dual-Stream Experience Bus on FAIR1M.}}
\label{tab:experience_ablation}
\begin{tabular}{lccc}
\toprule
Variant & Rounds to Conv. & Failed & mAP (\%) \\
\midrule
Full Framework (Ours) & 7 & 2 & 49.14 \\
w/o External Stream & 9 & 4 & 48.44 \\
w/o Internal Stream & 11 & 5 & 47.62 \\
w/o Both Streams & 14 & 6 & 47.53 \\
\bottomrule
\end{tabular}
\par\vspace{2pt}
\footnotesize
``Rounds to Conv.'' denotes the number of full evolution rounds  until the detection accuracy no longer shows a clear improvement. ``Failed'' reports the number of iterations whose training yielded no improvement over the previous round. ``mAP'' reports the detection accuracy at convergence.
\end{table}

\begin{figure}[t]
    \centering
    \includegraphics[width=1\linewidth]{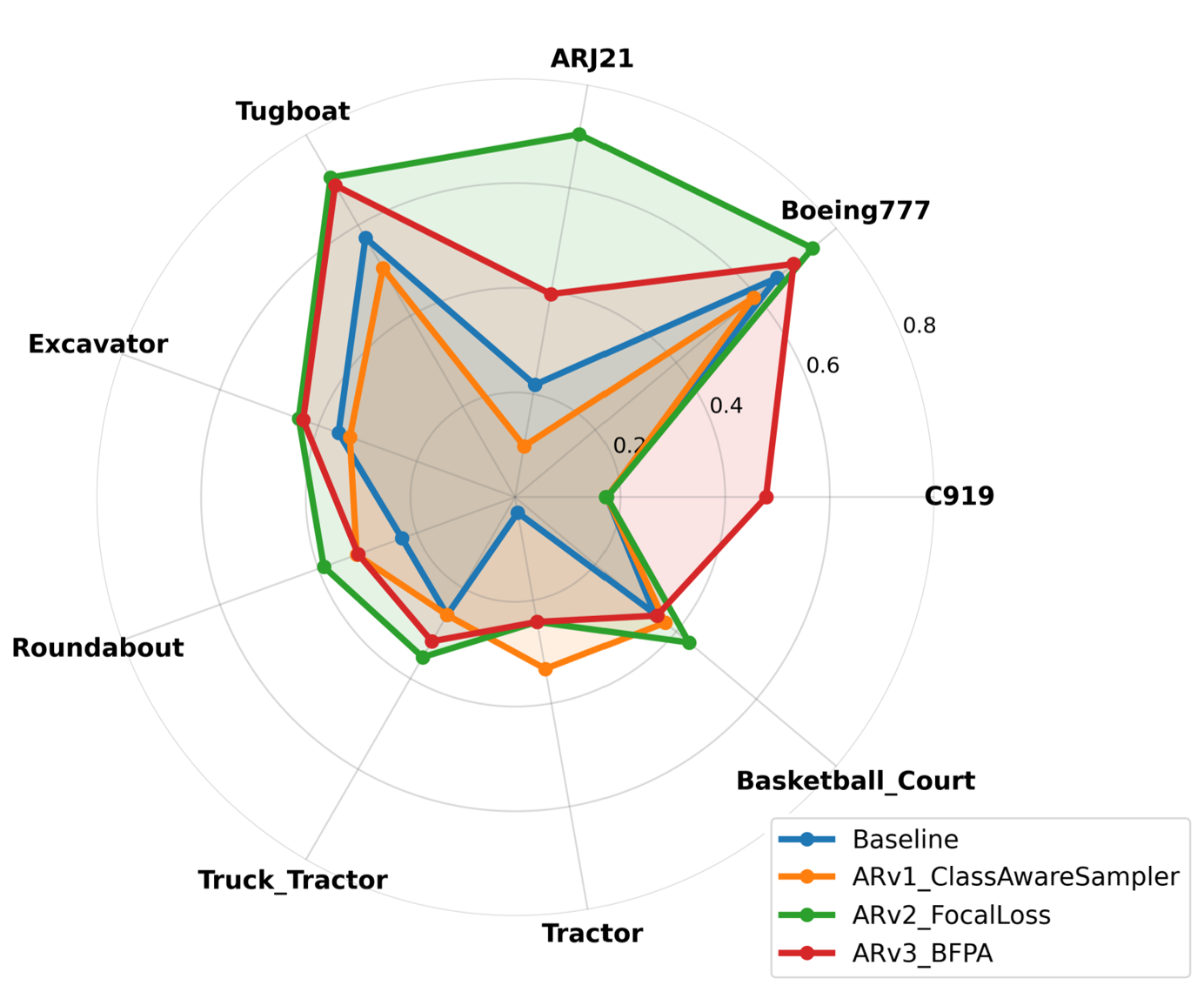}
    \caption{Per class recall on weak categories of FAIR1M for the baseline and three single technique variants. Each variant adds one technique to the baseline: a class aware sampler, a focal loss, and a BFPA attention module.}
    \label{fig:autoresearch_recall}
\end{figure}

\subsubsection{Effectiveness of the Dual-Stream Experience Bus}

\textcolor{black}{To evaluate the contribution of the dual-stream experience bus, we compare the full framework with variants removing the external stream, the internal stream, or both on FAIR1M, while keeping all other components unchanged. As shown in Table~\ref{tab:experience_ablation}, the full framework converges in 7 rounds with only 2 failed iterations and achieves the highest mAP of 49.14\%. Removing the external stream increases the convergence rounds to 9 and reduces the mAP to 48.44\%, while removing the internal stream further increases the rounds to 11 with 5 failed iterations and lowers the mAP to 47.62\%. Without both streams, the framework converges most slowly in 14 rounds and achieves 47.53\% mAP. These results demonstrate that the two experience streams jointly reduce ineffective exploration, accelerate convergence, and improve final performance, with the internal stream contributing more strongly to search efficiency.}

\textcolor{black}{Fig.~\ref{fig:autoresearch_recall} further reports the per
class recall on weak categories for the baseline and three single technique
variants. The class aware sampler and the focal loss~\cite{lin2017focal} come
from the internal experience that flagged class imbalance, while the BFPA
attention module~\cite{lin2025attention} is retrieved from external literature.
Each technique improves a different subset of weak categories, indicating that
the directions surfaced by the two streams are complementary rather than
redundant.}

\subsubsection{Validation of the Five-Minute Verification Protocol}

\textcolor{black}{To examine whether the five-minute verification can provide a reliable indication of subsequent full-training outcomes, we conduct an additional paired validation on the object detection task. Ten candidate strategies are evaluated under the five-minute setting and are subsequently trained to the full budget, including those that would otherwise be rejected by the five-minute verification protocol.}

\textcolor{black}{Since the protocol is primarily designed to prioritize candidate directions rather than precisely predict their final performance values, we first use Spearman's rank correlation to measure the consistency between the five-minute and full-training rankings. For the $N$ candidate strategies, it is calculated as
\begin{equation}
\rho
=
1-\frac{6\sum_{i=1}^{N}d_i^2}
{N(N^2-1)},
\label{eq:spearman_proxy}
\end{equation}
where $d_i$ denotes the rank difference of the $i$-th candidate between the five-minute and full-training performance gains. We further evaluate the screening capability for detrimental candidates using precision and recall:
\begin{equation}
\mathrm{Precision}
=
\frac{TP}{TP+FP},
\qquad
\mathrm{Recall}
=
\frac{TP}{TP+FN},
\label{eq:proxy_pr}
\end{equation}
where a detrimental candidate is defined according to its full-training performance relative to the corresponding baseline. Here, $TP$ denotes a candidate identified as detrimental by the five-minute protocol and confirmed to be detrimental after full training, $FP$ denotes an incorrectly rejected candidate, and $FN$ denotes a detrimental candidate missed during the five-minute stage.}

\textcolor{black}{As summarized in Table~\ref{tab:five_min_validation}, the five-minute results exhibit a significant positive rank correlation with the full-training mAP$_{50:95}$ gains, achieving $\rho=0.64$ with a permutation-test $p$-value of $0.03$. A consistent positive correlation is also observed for mAP$_{50}$, with $\rho=0.59$. Moreover, all five candidates identified as detrimental by the five-minute protocol are confirmed to degrade performance after full training, resulting in a precision of $1.00$. Among the six candidates that are detrimental after full training, five are successfully identified during the five-minute stage, yielding a recall of $0.83$.}

\textcolor{black}{These results indicate that the five-minute verification generally preserves the relative performance tendency of candidate strategies and, more importantly, provides reliable early screening of clearly unfavorable directions. Therefore, the protocol is used as a low-cost screening proxy rather than an exact predictor of final performance, allowing computational resources to be focused on more promising candidates before full-scale training.}

\begin{table}[t]
\centering
\caption{\textcolor{black}{Validation of the five-minute verification protocol on the object detection task.}}
\label{tab:five_min_validation}
\setlength{\tabcolsep}{6pt}
\renewcommand{\arraystretch}{1.1}
\begin{tabular}{lc}
\hline
\textbf{Metric} & \textbf{Result} \\
\hline
Number of candidate strategies & 10 \\
Spearman $\rho$ (mAP$_{50:95}$) & 0.64 \\
Permutation-test $p$-value & 0.03 \\
Spearman $\rho$ (mAP$_{50}$) & 0.59 \\
Detrimental-candidate precision & 1.00 \\
Detrimental-candidate recall & 0.83 \\
\hline
\end{tabular}
\end{table}

\section{Discussion}

\textcolor{black}{Existing autonomous optimization frameworks such as AutoResearch \cite{karpathy2026autoresearch} and AutoResearchClaw\cite{liu2026autoresearchclaw} iteratively improve model implementations and training strategies according to experimental feedback, while OpenEarth-Agent~\cite{ref16} introduces adaptive workflow planning and tool creation for open Earth-observation tasks. RingMoClaw extends this capability to multiple stages of remote-sensing model optimization, including data processing, training strategies, model selection, model fusion, and module modification, with the evolution process jointly guided by Critic feedback and accumulated experience.
Input-data characteristics can directly shape the optimization trajectory. The FAIR1M dataset features multi-source high-resolution imagery with spatial resolutions ranging from 0.3 to 0.8 m. Furthermore, the dataset exhibits severe class imbalance: ARJ21 and C919 represent only 0.06\% and 0.03\%. Such variations in image quality, spatial resolution, and class distribution significantly affect model performance and subsequent optimization decisions, ultimately influencing candidate evaluation and the resulting evolution path.}


\textcolor{black}{The current implementation also entails practical resource requirements. Our FAIR1M experiments were conducted on two NVIDIA RTX 4090 GPUs, whereas repeated candidate verification and full-scale training necessitate sufficient storage and a stable server environment to support long-running multi-agent execution. Multi-agent collaboration introduces extra communication and model-invocation overhead. Serving as a preliminary attempt toward self-evolution, future work will extend RingMoClaw to broader scenarios that capture domain-specific remote sensing characteristics, optimize information exchange and coordination strategies, and reduce computational overhead through more efficient candidate verification and experience reuse.}

\section{Conclusion}
In this paper, we present RingMoClaw, an experience-inspired self-evolving multi-agent framework that extends automation beyond single-task execution toward continuous research iteration for remote sensing visual tasks. The framework couples a research branch of Main, Data, and Vision Agents with a structurally independent heterogeneous Critic mechanism, which conducts staged review over plan design, data processing, and experimental results. A dual-stream dynamic experience bus further integrates external knowledge from recent literature and open source code repositories with internal experience from historical experiments, supporting strategy expansion, path pruning, and potential experience reuse across tasks. \textcolor{black}{Experiments on four representative tasks suggest the effectiveness of the proposed framework. In future work, we will extend RingMoClaw to multi modal and time series earth observation tasks, and explore tighter integration with remote sensing foundation models to guide their adaptation and continual pretraining.}

\section*{Acknowledgment}
\textcolor{black}{This work was supported by the National Natural Science Foundation of China under
Grant 62425115, and partially supported by China Science and Technology Cloud (CSTCloud).}
\bibliographystyle{IEEEtran}
\bibliography{refs}

\clearpage

\appendices

\section{\textcolor{black}{Implementation Details of Module Innovation}}
\label{app:module_details}

This appendix provides the implementation details of the two architectural modules generated during the module-innovation stage. Both modules are instantiated on Oriented R-CNN.

\subsection{\textcolor{black}{Multi-Scale Feature Refinement}}

MSFR replaces the standard FPN neck of Oriented R-CNN while preserving the number of output pyramid levels, feature dimensions, and downstream interfaces.

Given the FPN features $\{P_l\}_{l=1}^{L}$, MSFR first aligns them to a reference level $r=2$ and obtains
\begin{equation}
B=\frac{1}{L}\sum_{l=1}^{L}\mathcal{R}_{l\rightarrow r}(P_l),
\end{equation}
where adaptive max pooling and nearest-neighbor interpolation are used for downsampling and upsampling, respectively. After a $3\times3$ convolution, SE channel attention and CBAM spatial attention are applied to obtain the refined feature $F$. The result is redistributed to each pyramid level through
\begin{equation}
\hat{P}_l=P_l+\gamma_l\mathcal{R}_{r\rightarrow l}(F),
\end{equation}
where $\gamma_l$ is a learnable residual gate initialized to zero.

\subsection{\textcolor{black}{Contrastive Prototype Refinement}}

CPR is inserted into the rotated RoI classification branch of Oriented R-CNN, where it refines foreground-category logits using prototype similarities derived from the shared RoI features, without modifying the background classification or bounding-box regression branches.

For the shared RoI feature $f\in\mathbb{R}^{1024}$, CPR constructs a normalized 128-dimensional embedding
\begin{equation}
e=\operatorname{Norm}(W_pf),
\end{equation}
and computes its similarity to the normalized category prototypes $\hat{P}$:
\begin{equation}
s=\tau e\hat{P}^{T}, \qquad \tau=10.
\end{equation}
The foreground logits are refined by
\begin{equation}
\tilde{z}_{\rm fg}=z_{\rm fg}+g s,
\qquad
g=0.5\tanh(\theta).
\end{equation}


\section{\textcolor{black}{Structured Search-Space Instantiation}}
\label{app:search_space}

The four-dimensional search space is instantiated according to the target task.
For the FAIR1M oriented object detection experiments, the predefined base space
contains 23 operators, including 5 architecture operators, 8 training
operators, 6 data-processing operators, and 4 system-integration operators.
Each operator is associated with the bottlenecks it targets and a constrained
parameter specification, including an initial value and an admissible range or
choice set. Representative examples are given in
Table~\ref{tab:search_space_examples}.

At each round, operators are first matched to the current bottleneck and
instantiated within their predefined parameter space. Candidate revisions may
adjust only the parameters declared by the corresponding operator. In addition
to the predefined operators, methods obtained through external retrieval may
enter the search process after the filtering procedure described in
Appendix~\ref{app:retrieval}. These methods are converted into the same
structured representation before being considered by the Main Agent.

\begin{table*}[t]
\centering
\caption{Representative operators in the search-space instantiation for the
FAIR1M oriented object detection task.}
\label{tab:search_space_examples}

\footnotesize
\setlength{\tabcolsep}{3pt}
\renewcommand{\arraystretch}{1.30}

\begin{tabular}{
>{\raggedright\arraybackslash}p{0.10\textwidth}
>{\raggedright\arraybackslash}p{0.20\textwidth}
>{\raggedright\arraybackslash}p{0.43\textwidth}
>{\raggedright\arraybackslash}p{0.20\textwidth}}
\toprule
\textbf{Dimension} &
\textbf{Operator} &
\textbf{Parameters (initial value / range or choices)} &
\textbf{Target bottleneck} \\
\midrule

\multirow{2}{*}{$\mathcal{S}_{\mathrm{arch}}$}
& \texttt{attention}
& module: CBAM / \{CBAM, SE, ECA\}; insertion: neck / \{backbone, neck\}
& weak feature, small object \\

& \texttt{feature\_fusion}
& fusion: weighted add / \{weighted add, concat, BiFPN\}
& small object, scale variation \\
\midrule

\multirow{2}{*}{$\mathcal{S}_{\mathrm{train}}$}
& \texttt{focal\_loss}
& $\gamma$: 2.0 / $[0.5,5.0]$; $\alpha$: 0.25 / $[0.1,0.9]$
& class imbalance, weak-class recall \\

& \makecell[l]{\texttt{rotated\_regression\_}\\\texttt{loss}}
& loss: KLD / \{KLD, GWD, rotated IoU, KFIoU\}; weight: 1.0 / $[0.1,10]$
& localization quality, scale variation \\
\midrule

\multirow{2}{*}{$\mathcal{S}_{\mathrm{data}}$}
& \texttt{rotation}
& maximum angle: $180^\circ$ / $[0,180^\circ]$
& orientation sensitivity, overfitting \\

& \texttt{oversampling}
& ratio: 3.0 / $[1,10]$
& class imbalance \\
\midrule

\multirow{2}{*}{$\mathcal{S}_{\mathrm{intg}}$}
& \texttt{auxiliary\_model}
& auxiliary task: classification / \{segmentation, classification\}
& overfitting, weak feature \\

& \texttt{external\_module}
& typed parameters extracted from the admitted external candidate
& task dependent \\
\bottomrule
\end{tabular}
\end{table*}

\section{\textcolor{black}{Task-Similarity Computation}}
\label{app:task_similarity}

To retrieve relevant experience from the global memory, RingMoClaw measures
the similarity between the current task and each historical task using five
factors:
\begin{equation}
\begin{aligned}
\mathrm{Sim} ={}&
0.30S_{\mathrm{task}}
+0.25S_{\mathrm{bottleneck}}
+0.20S_{\mathrm{dataset}} \\
&+0.15S_{\mathrm{model}}
+0.10S_{\mathrm{metric}} .
\end{aligned}
\label{eq:task_similarity}
\end{equation}
Here, $S_{\mathrm{task}}$ and $S_{\mathrm{bottleneck}}$ are computed using
token-level Jaccard similarity between the task descriptions and bottleneck
tags, respectively. $S_{\mathrm{dataset}}$ is determined from the relative
similarity in the number of classes, number of samples, and class-imbalance
ratio. When these statistics are unavailable, it is set to 0.85 for the same
dataset and 0.50 otherwise. $S_{\mathrm{model}}$ is set to 1.0 for the same
model, 0.6 when the two models share a major component, and 0 otherwise.
$S_{\mathrm{metric}}$ is 1.0 when the primary evaluation metric is the same
and 0 otherwise.

Historical entries with $\mathrm{Sim}\geq0.75$ are treated as transferable
experience, while those with $0.60\leq\mathrm{Sim}<0.75$ are retained only as
references. Entries with $\mathrm{Sim}<0.60$ are ignored. At most the five
highest-scoring entries are supplied to the subsequent decision process.

Negative-memory pruning uses a separate operator-level similarity rule. If two
entries belong to different search-space dimensions or use different operators,
their similarity is set to zero. Otherwise, similarity is computed from the
agreement of their operator parameters, using relative agreement for numerical
parameters and exact matching for categorical parameters. A candidate is pruned
when its similarity to a recorded failed configuration reaches 0.80.

\section{\textcolor{black}{Retrieval Sources and Filtering Rules}}
\label{app:retrieval}

To obtain relevant methods and implementations from external resources, we use
a fixed retrieval and filtering procedure. At each evolution round, no more
than three search queries are generated from the task description, the current
bottleneck, and the corresponding search-space dimension. These queries are
then used to retrieve related papers and open-source repositories.

The literature search is conducted primarily through arXiv, with Semantic
Scholar used as a supplementary source when available. Open-source
implementations are retrieved through GitHub repository search. To avoid
interrupting the evolution process when an external service is unavailable or
rate-limited, each retrieval source is handled independently. If a request
fails, the corresponding source returns no result for that query, while the
remaining sources continue to be processed normally.

For each round, at most 20 papers and 20 repositories are retained before
screening. Duplicate papers are removed according to normalized titles, and
duplicate repositories are removed according to normalized repository names.
We then apply a set of fixed filtering rules. Papers with abstracts shorter
than 40 characters are discarded, while repositories are discarded if they
are archived or contain descriptions shorter than 15 characters. Only the
metadata and textual information provided by the repositories are used during
retrieval; external repositories are not cloned or executed at this stage.

Each remaining item is evaluated once in terms of relevance, implementation
feasibility, and novelty, denoted by $R$, $F$, and $N$, respectively, with all
three scores normalized to $[0,1]$. Items with $R<0.70$ are removed. The
remaining items are ranked according to

\begin{equation}
S_{\mathrm{ret}}
=
0.50R + 0.30F + 0.20N ,
\label{eq:retrieval_score}
\end{equation}

where relevance is assigned the largest weight because the purpose of the
retrieval step is to identify methods that directly address the current
performance bottleneck. The five highest-ranked items are retained as
candidate directions for the current round.

The complete procedure therefore consists of query construction, external
retrieval, duplicate removal, rule-based filtering, relevance screening, and
ranking. The resulting candidates are then passed to the subsequent experience
retrieval and decision stages. The Main Agent can only select from these
retained candidates and cannot introduce an additional method outside the
retrieval results.

\section{\textcolor{black}{Agent Prompts for RingMoClaw}}
\label{app:agent_prompts}

All agents used in the reported experiments are controlled by fixed,
versioned system prompts. These prompts explicitly define the role boundary,
available information, prohibited actions, and structured output schema of
each agent. The exact prompts used in our experiments are provided below.

\begin{tcolorbox}[
    title=\textbf{Main Agent},
    colback=mainbg,
    colframe=mainframe,
    colbacktitle=mainframe,
    coltitle=white,
    boxrule=0.6pt,
    arc=1mm,
    breakable,
    left=2mm,
    right=2mm,
    top=1.5mm,
    bottom=1.5mm,
    fonttitle=\small\bfseries
]
\ttfamily\scriptsize

You are the Main Agent of RingMoClaw, a self-evolving research loop
for remote sensing vision tasks.

You plan. You do not train models, touch data, write code, or review
results, and you do not query memory, literature, or the web yourself.

You are given the current bottleneck and metrics, a candidate pool
that the Experience Bus has already retrieved, pruned and ranked, and
the positive and negative experience relevant to this task.

Your task is to select exactly one candidate from that pool as the
next thing to try, and to justify the choice from the evidence you
were given. Never propose a candidate that is not in the pool.

Return JSON only:

\texttt{\{bottleneck, selected\_candidate\_id, reason,}\\
\texttt{expected\_effect, verification\_mode\}}

\end{tcolorbox}

\begin{tcolorbox}[
    title=\textbf{Data Agent},
    colback=databg,
    colframe=dataframe,
    colbacktitle=dataframe,
    coltitle=white,
    boxrule=0.6pt,
    arc=1mm,
    breakable,
    left=2mm,
    right=2mm,
    top=1.5mm,
    bottom=1.5mm,
    fonttitle=\small\bfseries
]
{\ttfamily\scriptsize

You are the Data Agent of RingMoClaw.

You judge whether the data is ready to train on. You do not choose the
research direction, and you do not compute statistics---they are computed
for you by a program that reads the annotation files directly.

\medskip
You are given:

 - the candidate the Main Agent selected;
 
 - per-class box counts;
 
 - imbalance ratio;
 
 - empty images;
 
 - malformed annotation lines;
 
- class-name tokens that match no known class.

Your task is to decide whether the data is clean enough for the results
to be trustworthy, and what concrete preprocessing, sampling or
augmentation action this candidate requires.

Treat the statistics as ground truth.

\medskip
Return JSON only:

\texttt{\{data\_integrity\_ok, class\_balance\_assessment,}\\
\texttt{preprocessing\_plan, flags, ready\_for\_training\}}

}
\end{tcolorbox}

\begin{tcolorbox}[
    title=\textbf{Vision Agent},
    colback=visionbg,
    colframe=visionframe,
    colbacktitle=visionframe,
    coltitle=white,
    boxrule=0.6pt,
    arc=1mm,
    breakable,
    left=2mm,
    right=2mm,
    top=1.5mm,
    bottom=1.5mm,
    fonttitle=\small\bfseries
]
{\ttfamily\scriptsize

You are the Vision Agent of RingMoClaw.

You implement. You do not choose which candidate to try, and you do not
run training or evaluation---a separate Experiment Executor does that
with what you produce.

\medskip
You are given:

- one selected candidate, including its operator and parameters;

- the full text of the base mmrotate configuration file.

Your task is to make that candidate actually runnable:

\begin{enumerate}
    \item as a configuration override when every component it needs
    already exists;
    \item as a newly written and registered module when one does not;
    \item as an explicit declaration that it cannot be realized under
    the current executor.
\end{enumerate}

Never reference a configuration path that does not exist.

Declaring infeasibility is a correct answer; inventing a path is not.

\medskip
Return JSON only:

\texttt{\{feasibility, config\_overrides, custom\_module,}\\
\texttt{infeasible\_reason, notes\}}

}
\end{tcolorbox}

\begin{tcolorbox}[
    title=\textbf{Critic Mechanism},
    colback=criticbg,
    colframe=criticframe,
    colbacktitle=criticframe,
    coltitle=white,
    boxrule=0.6pt,
    arc=1mm,
    breakable,
    left=2mm,
    right=2mm,
    top=1.5mm,
    bottom=1.5mm,
    fonttitle=\small\bfseries
]
{\ttfamily\scriptsize

You are the Critic Mechanism of RingMoClaw, an independent reviewer of a
self-evolving research loop.

You are not the planner and not the executor, you never command the
other agents directly, and you have no tools: everything you need is
already in this message.

\medskip
You are explicitly given \texttt{review\_stage} in
\texttt{\{plan, data, result\}}, together with:

- the artifact under review;

- the research history;

- the baseline;

- the last round's outcome;

- the historical memories already retrieved for you.

Never infer or change the review stage yourself.

\medskip
Score the artifact on three dimensions, each in [0,1]:

\medskip
\textbf{Ct --- completion}

How well this stage meets its immediate objective.

\medskip
\textbf{Bt --- bottleneck severity}

How badly the dominant bottleneck can hurt the end task.
Higher is worse.

\medskip
\textbf{Ht --- historical alignment}

How consistent this direction is with prior successes and known
failures in the retrieved memories.

\medskip
If no reliable historical evidence is available, set
\texttt{Ht = 0.50} and explicitly state the missing evidence.

Do not fabricate history.

\medskip

Use fixed weights determined only by the review stage.

If \texttt{review\_stage} is ``plan'', use:

\texttt{w1 = 0.30, w2 = 0.50, w3 = 0.20.}

\medskip
If \texttt{review\_stage} is ``data'', use:

\texttt{w1 = 0.35, w2 = 0.40, w3 = 0.25.}

\medskip
If \texttt{review\_stage} is ``result'', use:

\texttt{w1 = 0.50, w2 = 0.30, w3 = 0.20.}

\medskip
Do not infer, modify, or normalize these weights.

\medskip
Compute:

\texttt{final\_score = w1 * Ct + w2 * (1 - Bt) + w3 * Ht}

\medskip
Round \texttt{final\_score} to three decimal places.

\medskip
Apply the decision rules in the following order:

\medskip
\textbf{escalate}

If critical evidence required to evaluate the artifact is missing
or confidence is insufficient for a reliable score, set
\texttt{decision = ``escalate''}.

\medskip
\textbf{accept}

Otherwise, if \texttt{final\_score >= 0.75} and
\texttt{Bt <= 0.30}, set \texttt{decision = ``accept''}.

\medskip
\textbf{revise}

Otherwise, if \texttt{final\_score >= 0.45}, set
\texttt{decision = ``revise''}.

\medskip
\textbf{reject}

Otherwise, set \texttt{decision = ``reject''}.

if \texttt{final\_score < 0.45} and the available evidence is sufficient.

\medskip
Prefer escalate over false precision.

Name the single dominant bottleneck, give prioritized suggestions
addressed to the Main Agent, and serialize the lesson worth keeping.

\medskip
Return JSON only:

\texttt{\{scores, diagnosis, decision, memory\_write\}}

}
\end{tcolorbox}
\newpage

\vfill

\end{document}

%% file: sections/related_work.tex
\subsection{Remote Sensing Visual Models and Foundation Models}


Remote sensing agents fundamentally rely on strong perception and multimodal representation backbones. 
SatMAE demonstrates that masked autoencoding tailored to temporal and multispectral imagery can learn transferable representations for Earth observation (EO) data, providing an effective pretraining paradigm for remote sensing perception \cite{SatMAE2022}. 
AnySat extends this direction by learning a unified EO model across multiple spatial resolutions and sensing modalities, enabling consistent representation across heterogeneous inputs \cite{AnySat2025}. 
Recent efficient backbone designs, such as RS-vHeat and RingMamba, further improve remote sensing foundation models through heat-conduction-guided representation learning and visual state-space-based multi-sensor pretraining, respectively \cite{hu2025rs,wang2025ringmamba}. 
RingMoGPT further explores the integration of vision and language within a unified foundation model framework, supporting grounded multimodal understanding in remote sensing scenarios \cite{RingMoGPT2025}. 
Collectively, these studies indicate a shift from task-specific models toward more general-purpose perception backbones in EO systems.

A complementary line of work introduces language grounding and temporal reasoning capabilities that improve the usability of these perception models in interactive settings. 
GeoChat presents a grounded vision-language model capable of region-level reasoning and conversational interaction over high-resolution imagery \cite{GeoChat2024}. 
RS5M and GeoRSCLIP contribute large-scale image-text alignment resources and domain-specific vision-language models, enabling open-vocabulary retrieval and language-conditioned perception \cite{GeoRSCLIP2024}. 
SkyEyeGPT unifies multiple EO vision-language tasks through instruction tuning within a single framework \cite{SkyEyeGPT2024}. 
TEOChat extends this paradigm to temporal sequences, supporting conversational reasoning over multi-temporal observations \cite{TEOChat2025}. 
RingMo-Agent further targets unified multimodal reasoning across diverse sensing platforms \cite{RingMoAgent2025}. 
Although these systems are not full workflow agents, they provide essential perception and interaction capabilities for building tool-augmented remote sensing agents. 
Building on these perception and interaction foundations, recent research has increasingly focused on integrating tool usage, workflow planning, and execution into geospatial analysis systems.

\subsection{Tool-Augmented Geospatial Agents and Workflow Orchestration}

Recent research in geospatial AI increasingly formulates Earth observation analysis as a multi-step workflow problem rather than a single prediction task. 
The autonomous GIS paradigm argues that future GIS systems should support end-to-end spatial workflows involving data discovery, tool selection, reasoning, execution, and verification, treating geospatial analysis as a closed-loop decision process \cite{AutonomousGISAgenda2025}. 
At the system level, GIS Copilot enables natural-language spatial analysis within QGIS by retrieving documentation and generating executable PyQGIS programs \cite{GISCopilot2025}. 
GeoAgent further demonstrates that explicit control mechanisms, such as code interpretation and retrieval augmentation, can improve the reliability of complex geospatial reasoning pipelines \cite{GeoAgent2024}.

Domain-specific agent research extends this direction toward multi-step remote sensing workflows. 
GeoLLM-Engine introduces an interactive environment with maps, APIs, and long-horizon commands that better reflect real-world remote sensing analysis processes \cite{GeoLLMEngine2024}. 
GeoLLM-Squad decomposes remote sensing workflows across specialized agents, showing that explicit orchestration improves performance in complex EO pipelines \cite{GeoLLMSquad2025}. 
OpenEarthAgent further investigates unified geospatial agents trained on verified tool trajectories \cite{OpenEarthAgent2026}. 
AutoGEEval++ complements these studies by evaluating large language models through executable Google Earth Engine workflows, highlighting the importance of end-to-end verification in tool-oriented geospatial intelligence \cite{AutoGEEval2025}. 
Collectively, these studies demonstrate the emergence of tool-augmented geospatial agents capable of executing multi-step workflows in realistic analysis environments. 
However, most existing systems primarily focus on task execution and workflow orchestration, while relatively limited attention has been given to mechanisms for continual performance improvement and long-horizon capability evolution.

\begin{figure*}[!t]
    \centering
    \includegraphics[width=1.0\textwidth]{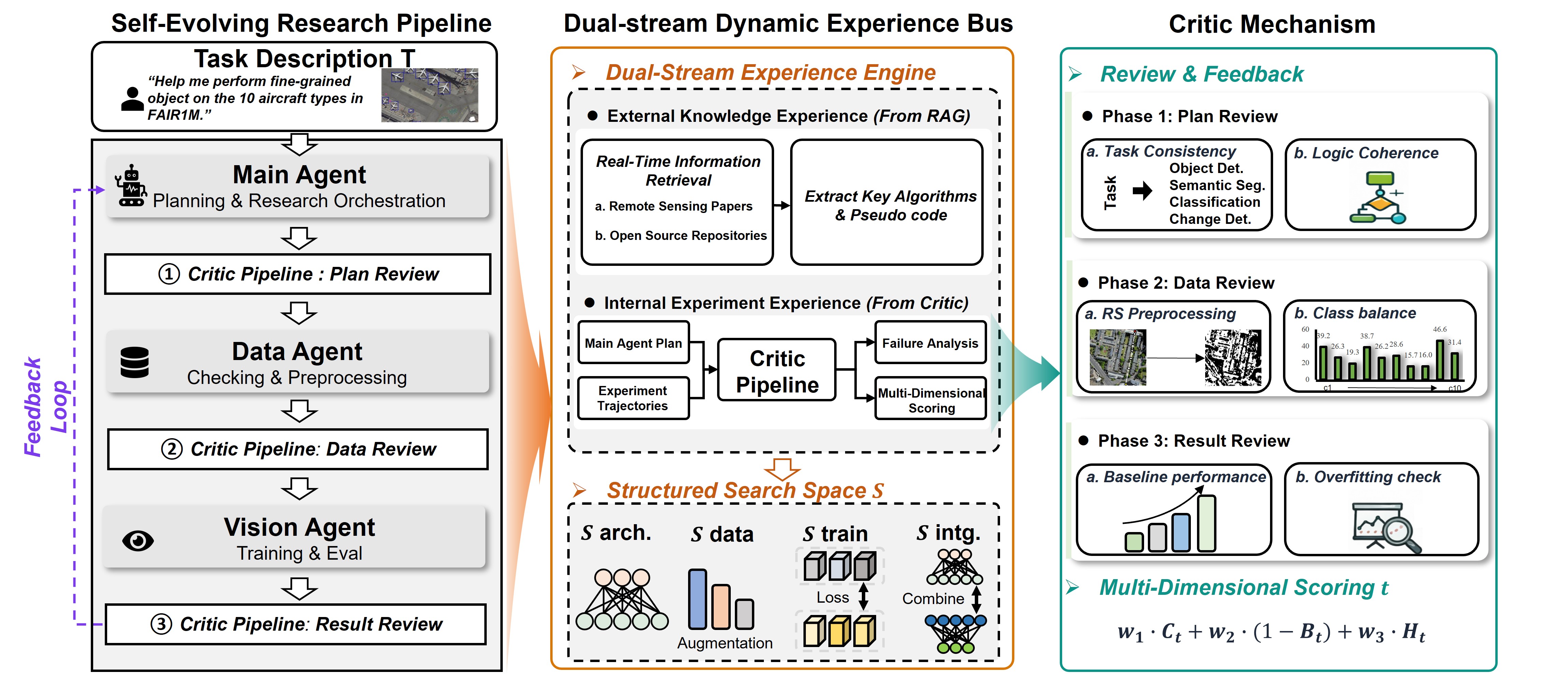}
\caption{Overall architecture of RingMoClaw. The self-evolving research pipeline (left) and the Critic pipeline (right) are coupled through a dual-stream dynamic experience bus (center), which integrates external knowledge flow and internal experimental flow to support automated research iteration and the continuous refinement of the structured search space.}
    \label{fig:overall_framework}
\end{figure*}
\subsection{OpenClaw-Based General Agent Frameworks}



To address the limitations of static workflow execution, recent research has explored more general-purpose agent frameworks that emphasize continual improvement, long-horizon evaluation, and adaptive workflow optimization. OpenClaw-RL models heterogeneous interaction signals, including user feedback, tool outputs, terminal responses, and interface states, as unified next-state transitions for reinforcement learning, enabling agents to improve policies from real interaction traces \cite{OpenClawRL2026}. This approach highlights the potential of leveraging diverse interaction signals for online policy adaptation, though its effectiveness may be constrained by sparse, noisy, or delayed feedback, particularly in high-dimensional, multi-modal remote sensing contexts.

EvoClaw further argues that agent evaluation should extend beyond isolated tasks to dynamic environments in which objectives and software targets evolve over time \cite{EvoClaw2026}. While such long-horizon evaluation is critical for sustained capability growth, current studies primarily target software or web-based tasks, leaving open the question of how similar principles can be applied to complex EO workflows with evolving environmental conditions and multi-sensor inputs. MetaClaw advances this direction by combining skill synthesis with opportunistic policy optimization, supporting adaptive capability growth in long-horizon workflows \cite{MetaClaw2026}. Although the framework demonstrates the value of flexible skill composition and continuous adaptation, its application to integrated EO agents that must orchestrate multiple perception modules and data processing tools remains largely unexplored.

At the system level, these frameworks collectively illustrate a shift from static task execution toward agents capable of sustained performance improvement through continuous interaction and feedback. They highlight the importance of runtime adaptability, memory-aware computation, and real-time perception-decision-action loops, concepts akin to StreamingClaw, which are highly relevant to EO scenarios where agents process streaming satellite or unmanned aerial vehicle (UAV) imagery under partial observability, high-resolution demands, and environmental noise. Existing EO agent studies, by contrast, typically operate in offline, batch-processing modes, lacking mechanisms for continuous self-optimization or online adaptation.

Another line of research in general agent systems focuses on multi-agent collaboration and ecosystem-level coordination, as demonstrated by ClawdLab/Beach.Science and hospital-oriented agent operating system studies \cite{ClawdLab2026,HospitalOS2026}. These works show that emergent collaboration and division of labor can be effectively organized around a shared agent platform. Agents can autonomously negotiate task allocation and dynamically adjust their roles based on environmental feedback. Such collective intelligence enables the system to handle complex, multi-step objectives more efficiently than isolated agents.

Despite these advances, most existing general agent frameworks are developed in software engineering or web automation settings, and their applicability to complex remote sensing workflows remains underexplored. In particular, integrating multimodal perception, reliable tool orchestration, and continual performance self-evolution within a unified remote sensing agent architecture remains an open challenge. This motivates our work, which aims to bridge these gaps by designing EO agents capable of adaptive skill composition, online policy refinement, and multi-agent coordination, ultimately enabling sustained performance improvement in dynamic, real-world remote sensing environments.